\RequirePackage{fix-cm}
\documentclass[twocolumn]{svjour3}    
\pdfoutput=1

\usepackage{graphicx}
\usepackage{natbib}

\usepackage{url}            
\usepackage{booktabs}       
\usepackage{amsfonts}       
\usepackage{nicefrac}       
\usepackage{microtype}      
\usepackage{graphicx}
\usepackage{enumitem}
\usepackage{bm}
\usepackage{nccmath}
\usepackage{boldline}
\usepackage{booktabs} 

\usepackage{algorithm}
\usepackage[noend]{algcompatible}

\usepackage{capt-of}
\usepackage{color,soul}
\usepackage[pagebackref=false,breaklinks=true,colorlinks,bookmarks=false,urlcolor=blue, citecolor=blue]{hyperref}

\usepackage{times}
\usepackage{epsfig}
\usepackage{multirow}
\usepackage{xcolor}
\usepackage{amsmath}
\usepackage{amssymb}
\usepackage{wrapfig}
\usepackage{comment}

\usepackage{textcomp}
\usepackage{bm}
\usepackage{soul}

\usepackage{makecell}
\usepackage{subfigure}
\usepackage{arydshln}
\usepackage{ulem}

\usepackage{tabularx}
\usepackage{tabularx,booktabs}
\newcolumntype{C}{>{\centering\arraybackslash}X}
\usepackage{cuted}
\usepackage{capt-of}
\usepackage{adjustbox}
\usepackage{multirow}

\newcommand{\etal}{\textit{et al}. }

\useunder{\underline}{\ul}{}

\usepackage{tikz}

\begin{document}

\title{DepthFormer: Exploiting Long-Range Correlation and Local Information \\ for Accurate Monocular Depth Estimation}

\author{Zhenyu Li \and Zehui Chen \and Xianming Liu \and  Junjun Jiang}

\institute{Zhenyu Li \and Xianming Liu \and Junjun Jiang \at
         School of Computer Science and Technology,\\ 
         Harbin Institute of Technology, \\ 
         \email{{zhenyuli17, csxm, jiangjunjun}@hit.edu.cn}
         \and
         Zehui Chen \at
         Department of Automation, \\
         University of Science and Technology of China, \\
         \email{lovesnow@mail.ustc.edu.cn}
}

\date{Received: date / Accepted: date}

\maketitle

\begin{abstract}
   This paper aims to address the problem of supervised monocular depth estimation. We start with a meticulous pilot study to demonstrate that the long-range correlation is essential for accurate depth estimation. Therefore, we propose to leverage the Transformer to model this global context with an effective attention mechanism. We also adopt an additional convolution branch to preserve the local information as the Transformer lacks the spatial inductive bias in modeling such contents. However, independent branches lead to a shortage of connections between features. To bridge this gap, we design a hierarchical aggregation and heterogeneous interaction module to enhance the Transformer features via element-wise interaction and model the affinity between the Transformer and the CNN features in a set-to-set translation manner. Due to the unbearable memory cost caused by global attention on high-resolution feature maps, we introduce the deformable scheme to reduce the complexity. Extensive experiments on the KITTI, NYU, and SUN RGB-D datasets demonstrate that our proposed model, termed DepthFormer, surpasses state-of-the-art monocular depth estimation methods with prominent margins. Notably, it achieves the most competitive result on the highly competitive KITTI depth estimation benchmark. Our codes and models are available\footnote{\url{https://github.com/zhyever/Monocular-Depth-Estimation-Toolbox}}.

\end{abstract}
\section{Introduction}
\label{sec:introduction}

Monocular depth estimation plays a critical role in three dimensional reconstruction and perception. Since the groundbreaking work of~\citep{he2016resenet}, convolutional neural network (CNN) has dominated the primary workhorse for depth estimation, in which the encoder-decoder based architecture is designed~\citep{fu2018deep, lee2019bts, bhat2021adabins}. Although there have been numerous work focusing on the decoder design~\citep{fu2018deep, bhat2021adabins}, recent studies suggest that the encoder is even more pivotal for accurate depth estimation~\citep{lee2019bts, ranftl2021dpt}. Due to the lack of depth cues, fully exploiting both the long-range correlation (\textit{i.e.}, distance relationship among objects) and the local information (\textit{i.e.}, consistency of the same object) are critical capabilities of an effective encoder~\citep{saxena2005learning}. Therefore, the potential bottleneck of current depth estimation methods may lie in the encoder where the convolution operators can scarcely model the long-range correlation with a limited receptive field~\citep{ranftl2021dpt}.

\begin{figure}
    \centering
    \includegraphics[width=3.2in]{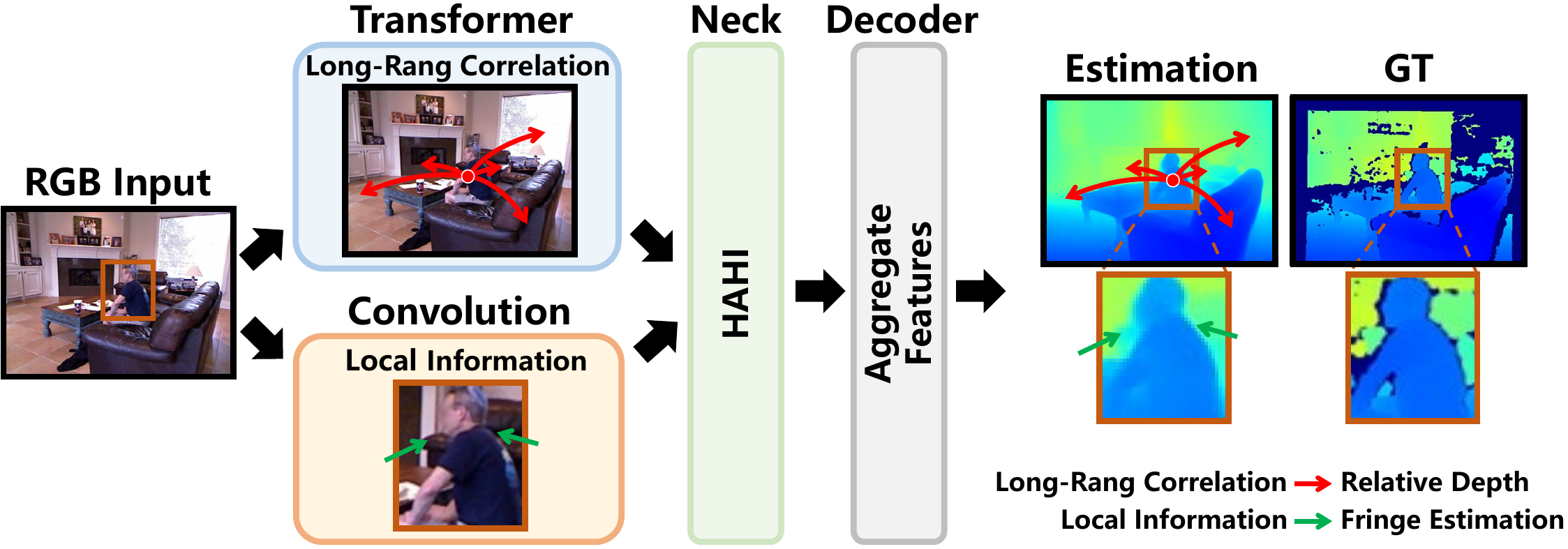}
    \caption{Overview: We design an encoder consisting of a Transformer branch to learn the long-range correlation and a convolution branch to extract the local information. To alleviate the lack of connections between the two branches, we propose the HAHI module to enhance features and model affinities.}
    \label{fig:teaser}
\end{figure}

\begin{figure*}[t]
\centering
  \includegraphics[width=1\linewidth]{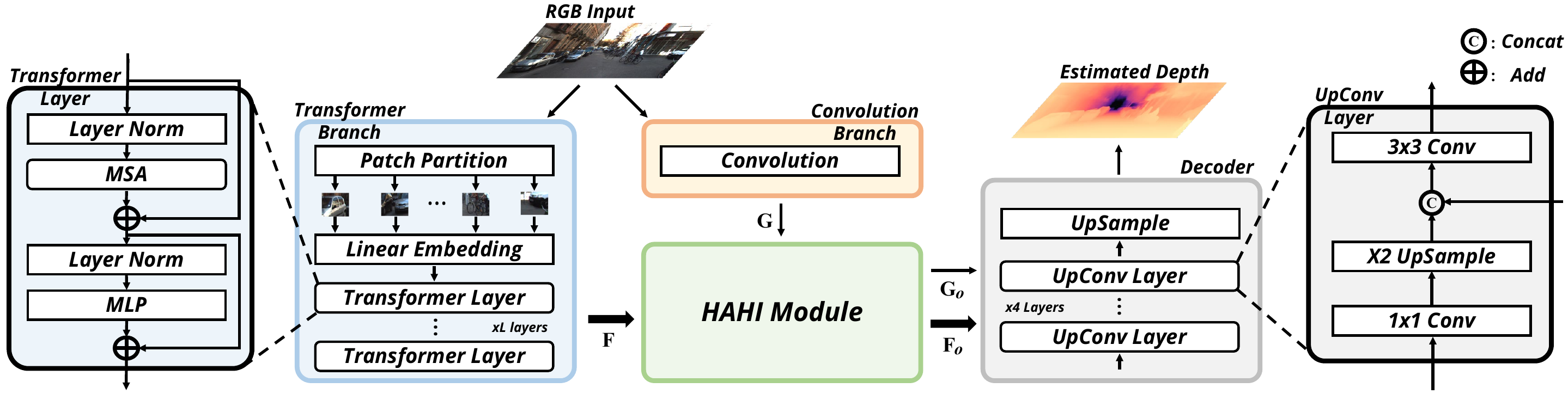}
  \caption{An overview of DepthFormer. It comprises three major components: an encoder consisting of a Transformer branch and a convolution branch, a hierarchical aggregation and heterogeneous interaction (HAHI) module, and a standard decoder. The HAHI enhances the Transformer features ${\mathbf{F}}$ and models the affinity between the Transformer and the convolution features $\mathbf{G}$.}
\label{fig:arch}
\end{figure*}

In terms of CNN, there have been great efforts to overcome the above limitation, roughly grouped into two categories: manipulating the convolution operation and integrating the attention mechanism. The former applies  advanced variations, including multi-scale fusion~\citep{ronneberger2015u}, atrous convolutions~\citep{chen2017deeplab} and feature pyramids~\citep{zhao2017pyramid}, to improve the effectiveness of convolution operators. The latter introduces the attention module~\citep{vaswani2017transformer} to model the global interactions of all pixels in the feature map. There are also several general approaches~\citep{fu2018deep, lee2019bts, huynh2020guiding, bhat2021adabins} that explore the combination of both these strategies. Though the performance is improved significantly, the dilemma persists.

In an alternative to CNN, Vision Transformer (ViT)~\citep{dosovitskiy2020vit}, which achieves tremendous success on image recognition, demonstrates the advantages of serving as the encoder for depth estimation. Benefiting from the attention mechanism, the Transformer is more expert at modeling the long-range correlation with a global receptive field. However, our pilot study (Sec.~\ref{sec:subsec:motivation}) indicates the ViT encoder cannot produce satisfactory performance due to the lack of spatial inductive bias in modeling the local information~\citep{yang2021transdepth}.

To mitigate these issues, we propose a novel monocular depth estimation framework, \textbf{DepthFormer} (illustrated in Fig.~\ref{fig:teaser}), which boosts model performance by incorporating the advantages from both the Transformer and the CNN. The principle of DepthFormer lies in the fact that the Transformer branch models the long-range correlation while the additional convolution branch preserves the local information. We argue that the integration of these two-type features can help achieve more accurate depth estimation. However, independent branches with late fusion lead to insufficient feature aggregation for the decoder. To bridge this gap, we design the Hierarchical Aggregation and Heterogeneous Interaction (HAHI) module to combine the best part of both branches. Specifically, it consists of a self-attention module to enhance the features among hierarchical layers of the Transformer branch via element-wise interaction and a cross-attention module to model the affinity between `heterogeneous' features (\textit{i.e.}, Transformer and CNN features) in a set-to-set translation manner. Since global attention on high-resolution feature maps leads to an unbearable memory cost, we propose to leverage the deformable scheme~\citep{dai2017deformable,zhu2020deformabledetr} that only attends to a limited set of key sampling vectors in a learnable manner to alleviate this problem.

The main contributions of this work are three-fold: (1) We apply the Transformer as the image encoder to exploit the long-range correlation and adopt an additional convolution branch to preserve the local information. (2) We design the HAHI to enhance features via element-wise interaction and model the affinity in a set-to-set translation manner. (3) Our proposed approach DepthFormer significantly outperforms state-of-the-arts with prominent margins on the KITTI~\citep{geiger2013kitti}, NYU~\citep{silberman2012nyu} and SUN RGB-D~\citep{song2015sun} datasets. Furthermore, it achieves the most competitive result on the highly competitive KITTI depth estimation benchmark\footnote{\url{http://www.cvlibs.net/datasets/kitti/eval_depth.php?benchmark=depth_prediction}}.

\begin{figure*}[t]
    \centering
    \footnotesize
    \begin{tabular}{l}
        \includegraphics[width=0.995\linewidth]{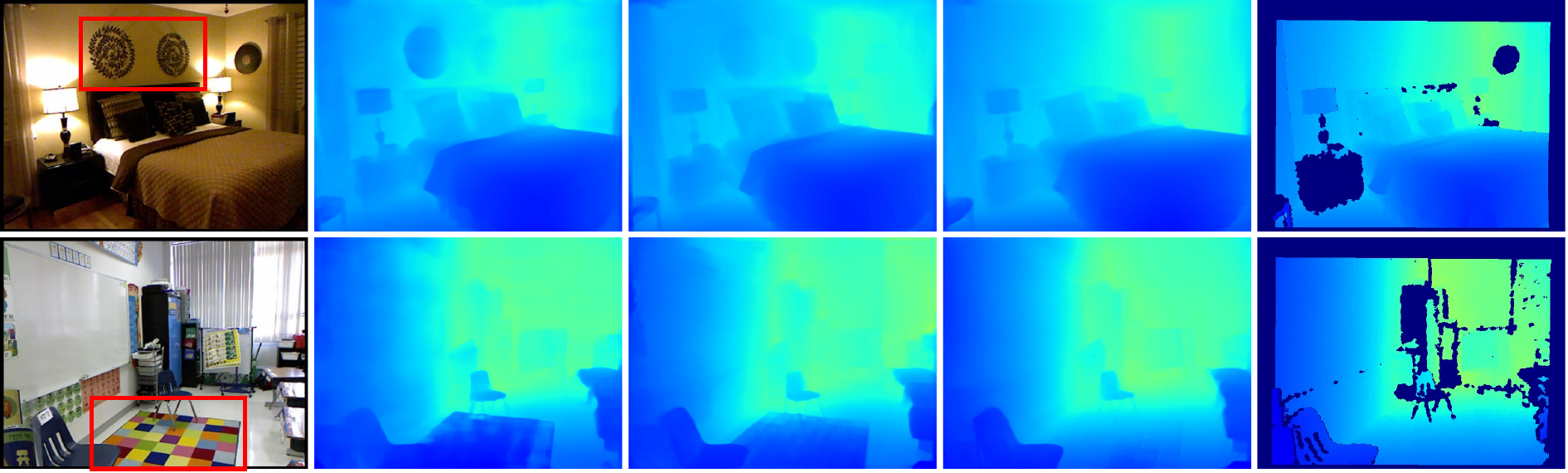} \\
        \hspace{0.08\linewidth}RGB
        \hspace{0.09\linewidth}BTS~\citep{lee2019bts}
        \hspace{0.02\linewidth}Adabins~\citep{bhat2021adabins}
        \hspace{0.07\linewidth}Ours
        \hspace{0.16\linewidth}GT\\
    \end{tabular}
    \caption{Failure cases of previous methods on NYU dataset caused by a limited receptive field of the convolution operator.}
    \label{fig:pilot-study-sota}
 \end{figure*}

\section{Related Work}
\label{sec:related_work}

Estimating depth from RGB images is an ill-posed problem. Lack of cues, scale ambiguities, translucent or reflective materials all leads to ambiguous cases where appearance cannot infer the spatial construction. With the rapid development of deep learning, CNN has become a key component of mainstream methods to provide reasonable depth maps from a single RGB input.

\textbf{Monocular depth estimation} has drawn much attention in recent years. Among  numerous effective methods, we consider DPT \citep{ranftl2021dpt}, Adabins \citep{bhat2021adabins} and Transdepth \citep{yang2021transdepth} as three the most important competitors.

DPT proposes to utilize ViT as the encoder and pre-train models on larger-scale depth estimation datasets. Adabins uses adaptive bins that dynamically change depending on representations of the input scene and proposes to embed the mini-ViT at a high resolution (after the decoder). Transdepth embeds ViT at the bottleneck to avoid the Transformer losing the local information and presents an attention gate decoder to fuse multi-level features. We focus on comparing these (and many other) methods in this paper.

\textbf{Encoder-decoder} is commonly used in monocular depth estimation \citep{eigen2014depth, fu2018deep, hu2019revisiting, lee2019bts, huynh2020guiding, yang2021transdepth, bhat2021adabins}. In terms of the encoder, mainstream feature extractors, including EfficientNet~\citep{tan2019efficientnet}, ResNet~\citep{he2016resenet} and DenseNet~\citep{huang2017densenet}, are adopted to learn representations. The decoder frequently consists of successive convolutions and upsampling operators to aggregate encoder features in a late fusion manner, recover the spatial resolution and estimate the depth. In this paper, we utilize the baseline decoder architecture in~\citep{alhashim2018densedepth}. It allows us to more explicitly study the performance attribution of key contributions of this work, which are independent of the decoder.

\textbf{Neck} modules between the encoder and the decoder are proposed to enhance features. Many previous methods only focus on the bottleneck feature but ignore the lower-level ones, limiting the effectiveness~\citep{fu2018deep, lee2019bts, huynh2020guiding, yang2021transdepth}. In this work, we propose the HAHI module to enhance all the multi-level hierarchical features. When another branch is available, it can model the affinity between the two-branch features as well, which benefits the decoder to aggregate the heterogeneous information.

\textbf{Transformer} networks are gaining greater interest in the computer vision community~\citep{dosovitskiy2020vit, liu2021swin, carion2020detr, zheng2021setr}. Following the success of recent trends that apply the Transformer to solve computer vision tasks, we propose to leverage the Transformer as the encoder to model long-range correlations. In Sec.~\ref{sec:subsec:motivation}, we discuss our motivation and present differences between our method and several related work~\citep{ranftl2021dpt, yang2021transdepth, bhat2021adabins} that adopt the Transformer in monocular depth estimation.

\begin{table*}[t]
    \centering
        \begin{adjustbox}{width=0.8\linewidth,center}
            \begin{tabular}{@{}llccccc@{}}
            \toprule
            Backbone  & Range & \textbf{$\delta_1$}$\uparrow$ & \textbf{$\delta_2$}$\uparrow$ & \textbf{$\delta_2$}$\uparrow$ & REL$\downarrow$ & RMS$\downarrow$ \\ 
            \midrule
            \multirow{3}{*}{ResNet-50~\citep{he2016resenet}}  
            & 0$m$-20$m$ & 0.973  & 0.998 & 1 & 0.054 &  0.985  \\
            & 60$m$-80$m$ & 0.600 & 0.900 & 0.972  & 0.188 & 14.30  \\
            & Overall & 0.952 & 0.994 & 0.999 & 0.065 &  2.596 \\
            \midrule
            \multirow{3}{*}{ViT-Base~\citep{dosovitskiy2020vit}} 
            & 0$m$-20$m$ & 0.955 & 0.995 & 0.999 & 0.071 & 1.275   \\
            & 60$m$-80$m$ & 0.727 & 0.936 & 0.985 & 0.150 & 11.86  \\
            & Overall & 0.938 & 0.992 & 0.999 & 0.080 & 2.695  \\
            \bottomrule
            \end{tabular}
        \end{adjustbox}
    \caption{Pilot study results on the KITTI dataset. Overall means the measurements are made from 0$m$ to 80$m$.}
    \label{tab:kitti_pilot}
\end{table*}

\begin{figure*}[t]
    \centering
    \footnotesize
    \begin{tabular}{@{}c@{\hspace{0.06cm}}c@{\hspace{0.06cm}}c@{\hspace{0.06cm}}c@{}}
        \includegraphics[width=0.30\linewidth]{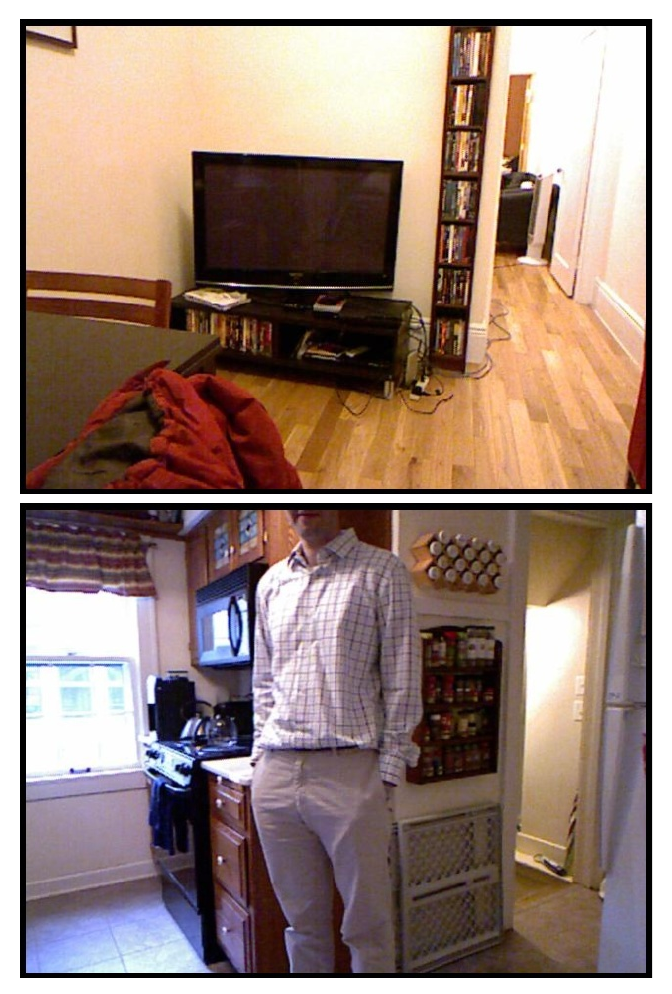} & 
        \includegraphics[width=0.30\linewidth]{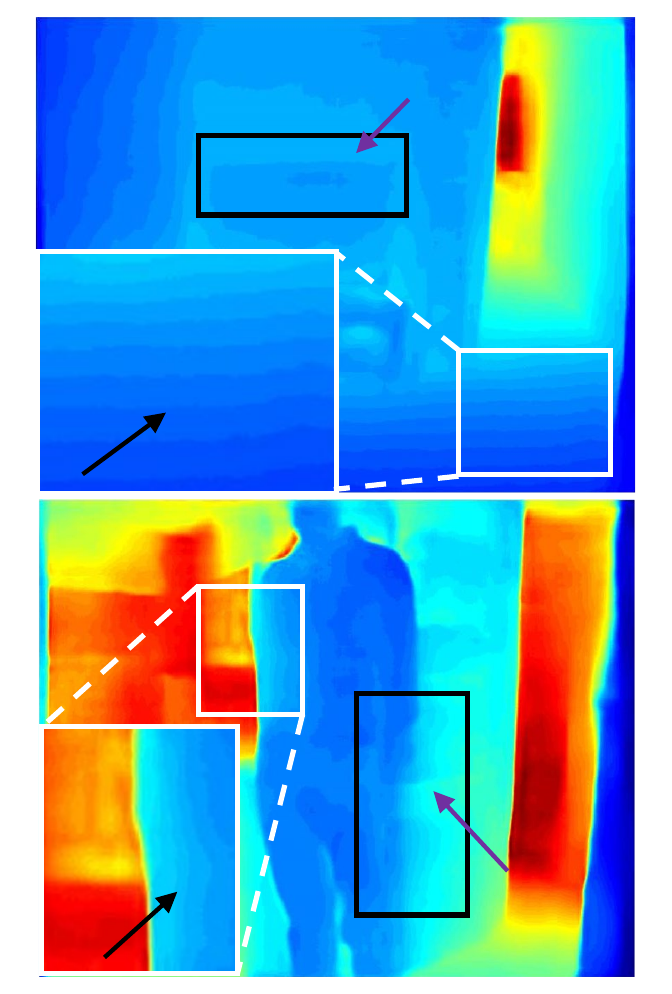} & 
        \includegraphics[width=0.30\linewidth]{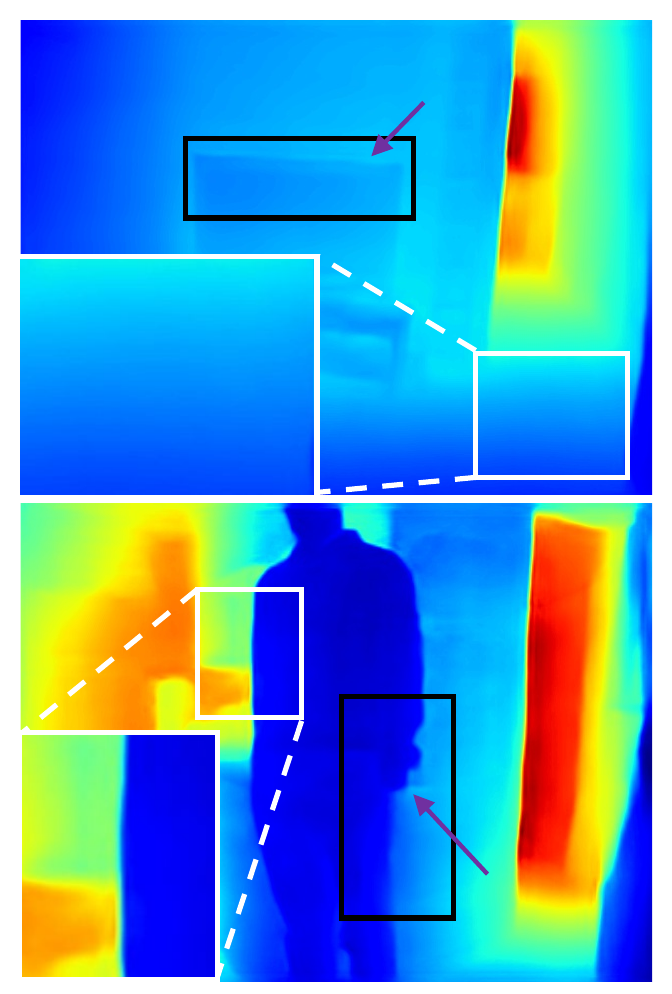} \\
        RGB & TransDepth~\citep{yang2021transdepth} & Ours\\
    \end{tabular}
    \caption{Demonstration of artifacts and the lost of local information. Our method provides consistent and sharp depth estimation.}
    \label{fig:compare-Transdepth}
 \end{figure*}

\section{Methodology}
\label{sec:methodology}

In this section, we present the motivation of this work and introduce the key components of DepthFormer: (1) an encoder consisting of a Transformer branch and a convolution branch and (2) the hierarchical aggregation and heterogeneous interaction (HAHI) module. An overview of DepthFormer is shown in Fig.~\ref{fig:arch}.

\subsection{Motivation}
\label{sec:subsec:motivation}
To indicate the necessity of this work, we conduct a meticulous pilot study to investigate the limitations of existing methods that utilize pure CNN or ViT as the encoder in monocular depth estimation.

\textbf{Pilot Study}: We first present several failure cases of the state-of-the-art CNN-based monocular depth estimation methods on the NYU dataset in Fig.~\ref{fig:pilot-study-sota}. The depth results at the wall decorations and carpets are unexpectedly incorrect. Due to the pure convolutional encoder for feature extraction, it is hard for them to model the global context and capture the long-range distance relationship among objects through limited receptive fields. Such large-area counter-intuitive failures severely impair the model performance.

To solve the above issue, ViT can serve as a proper alternative that is superior in modeling the long-range correlation with a global receptive field. Therefore, we experiment to analyze the performance of ViT- and CNN-based methods on the KITTI dataset. Specifically, based on DPT~\citep{ranftl2021dpt}, we adopt ViT-Base and ResNet-50 as the encoder to extract features, respectively. The results shown in Tab.~\ref{tab:kitti_pilot} prove that the models applying ViT as encoder outperform those using ResNet-50 on distant object depth estimation. However, opposite results appear on the near objects. Since the depth values exhibit a long tail distribution and there are much more near objects in scenes~\citep{jiao2018look}, the overall results of the models applying ViT are significantly inferior. More experimental details and results are reported in the experimental section.

\textbf{Analysis}: In general, it is tougher to estimate the depth of distant objects directly. Benefiting from modeling the long-range correlation, the ViT-based model can be more reliable to accomplish it via reference pixels in a global context. The knowledge of distance relationships among objects results in better performance on distant object depth estimations. As for the inferior near object depth estimation result, there are many potential explanations. We highlight two major concerns: (1) The Transformer lacks spatial inductive bias in modeling the local information \citep{yang2021transdepth}. As for depth estimation, the local information is reflected in the detailed context that is crucial for consistent and sharp estimation results. However, these detailed content tends to be lost during the patch-wise interaction of the Transformer. Since objects appearing nearer are larger with higher texture quality~\citep{saxena2005learning}, the Transformer will lose more details at these locations, which severely deteriorates the model performance at a near range and leads to unsatisfying results. (2) Visual elements vary substantially in scale~\citep{liu2021swin}. In general, a U-Net~\citep{ronneberger2015u} shape architecture is applied for depth estimation, where the multi-scale skip connections are pivotal for exploiting multi-level information. Since the tokens in ViT are all of a fixed scale, the consecutive non-hierarchical forward propagation makes the multi-scale property ambiguous, which may also limit the performance.

In this paper, we propose to leverage an encoder consisting of Transformer and convolution branches to exploit both the long-range correlation and local information. Different from DPT~\citep{ranftl2021dpt}, which directly utilizes ViT as the encoder, we introduce a convolution branch to make up for the deficiencies of spatial inductive bias in the Transformer branch. Furthermore, we replace ViT with Swin Transformer~\citep{liu2021swin} so that the Transformer encoder can provide hierarchical features and reduce the computational complexity. Unlike previous methods which embed the Transformer into the CNN~\citep{bhat2021adabins, yang2021transdepth}, we adopt the Transformer to encode images directly, which can fully exert the advantages of the Transformer and avoid the CNN discarding crucial information before the global context modeling. Moreover, due to the independence of these two branches, the simple late fusion of the decoder leads to an insufficient feature aggregation and marginal performance improvement. To bridge this gap, we design the HAHI module to enhance features and model affinities via feature interaction, which alleviates the deficiency and helps combine the best part of both branches.

\begin{figure*}[t]
    \centering
      \includegraphics[width=1\linewidth]{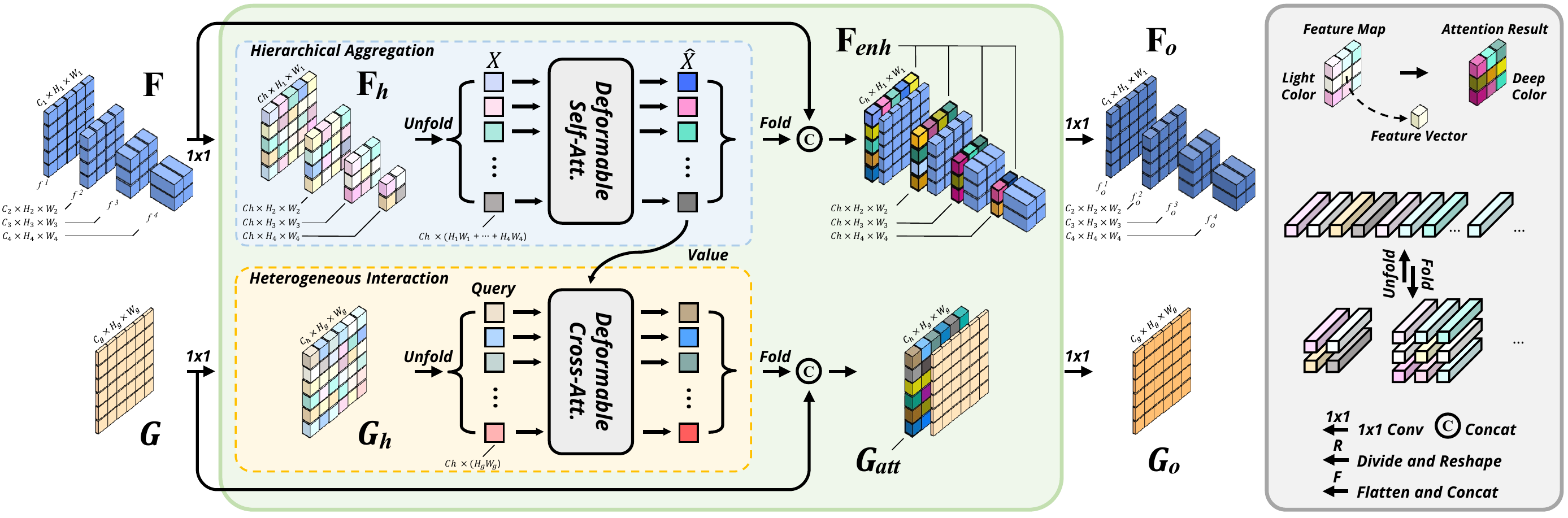}
      \caption{Illustration of our proposed HAHI. The deformable self-attention module enhances the input Transformer features $\mathbf{F}$. The deformable cross-attention module models the affinity between the Transformer features $\mathbf{F}$ and the convolution features $\mathbf{G}$ in a set-to-set translation manner. The output of the HAHI, ${\mathbf{F}}_o$ and $\mathbf{G}_o$, are sent to the decoder for the final aggregation. Due to the adoption of Swin Transformer layers to extract the hierarchical features, $\mathbf{F}$ exhibits different sizes and channels.}
    \label{fig:HAHI}
\end{figure*}

\subsection{Transformer and CNN Feature Extraction}
\label{sec:subsec:FE}
We propose to extract image features via an encoder consisting of a Transformer branch and a light-weight convolution branch, thus fully exploiting the long-range correlation and the local information. 

\textbf{Transformer branch} first splits the input image $I$ into non-overlapping patches by a patch partition module. The initial feature representation of each patch is set as a concatenation of the pixel RGB values. After that, a linear embedding layer is applied to project the initial feature representation to an arbitrary dimension, which is served as the input of the first Transformer layer and denoted as ${\mathbf{z}}^0$. After that, $L$ Transformer layers are applied to extract features. In general, each layer consists of a multi-head self-attention (MSA) module, followed by a multi-layer perceptron (MLP). A LayerNorm (LN) is applied before the MSA and the MLP, and a residual connection is utilized for each module. Therefore, the process of layer $l$ is formulated as
\begin{align}
   &{\hat{\mathbf{z}}^{l}} = \text{MSA}\left( {\text{LN}\left( {{{\mathbf{z}}^{l - 1}}} \right)} \right) + {\mathbf{z}}^{l - 1},\nonumber\\
   &{\mathbf{z}^l} = \text{MLP}\left( {\text{LN}\left( {{{\hat{\mathbf{z}}}^{l}}} \right)} \right) + {{\hat{\mathbf{z}}}^{l}},\label{eq.swin}
\end{align}
where ${\hat{\mathbf{z}}}^l$ and ${\mathbf{z}}^l$ denote the output features of the MSA module and the MLP module for layer $l$, respectively. The structure of a Transformer layer is illustrated in the left part of Fig.~\ref{fig:arch}.  Following DPT~\citep{ranftl2021dpt}, we sample and reassemble $N$ feature maps from the $N$ selected Transformer layers as the output of the Transformer branch and symbolize them as ${\mathbf{F}}=\{f^n\}_{n=1}^N$, where $f^n\in \mathbb{R}^{C_n\times H_n\times W_n}$ indicates the $n^{th}$ reassembled feature map.

Notably, our framework is compatible with a variety of Transformer structures. In this paper, we prefer to utilize Swin Transformer~\citep{liu2021swin} to provide hierarchical representations and reduce the computational complexity. The main differences from the standard Transformer layers lie in the local attention mechanism, the shifted window scheme, and the patch merging strategy.

\textbf{Convolution branch} contains a standard ResNet encoder to extract the local information, which is commonly used in depth estimation methods. Only the \textit{first} block of the ResNet is used here to exploit the local information, which avoids the low-level features being washed out by consecutive multiplications~\citep{yang2021transdepth} and greatly reduces the computational time. The output feature map with $C_g$ channels is denoted as $\mathbf{G}\in \mathbb{R}^{C_g\times H_g\times W_g}$.

Upon acquiring Transformer features $\mathbf{F}$ and convolution features $\mathbf{G}$, we feed them to the HAHI module for further processing. Compared to TransDepth~\citep{yang2021transdepth}, we adopt an additional convolution branch to preserve the local information. It avoids the discarding of crucial information by CNN and enables us to predict sharper depth maps without artifacts, as shown in Fig.~\ref{fig:compare-Transdepth}.

\subsection{HAHI Module}
\label{sec:subsec:HAHI}

To alleviate the limitation of insufficient aggregation, we introduce the HAHI module to enhance the Transformer features and further model the affinity of the Transformer and the CNN features in a set-to-set translation manner. It is motivated by Deform-DETR~\cite{zhu2020deformabledetr} and attempt to apply attention modules to solve the fusion of heterogeneous features.

We consider a set of hierarchical features $\textbf{F}=\{{f^n}\}_{n=1}^{N}$ as the inputs for feature enhancement. Since we use the Swin Transformer layers to extract the features, the reassembled feature maps will exhibit different sizes and channels, as shown in Fig.~\ref{fig:HAHI}. Many previous works have to downsample the multi-level features to the resolution of the bottleneck feature and can only enhance the bottleneck feature with simple concatenation, or latent kernel schemes~\citep{yang2021transdepth, zheng2021setr, lee2019bts}. Oppositely, we aim to enhance all the features without downsampling operators that may lead to information loss.

Specifically, we first utilize 1$\times$1 convolutions to project all the hierarchical features to the same channel $C_h$, denoting as ${\mathbf{F}}_h=\{{f_h^n}\}_{n=1}^{N}$. Then, we unfold (\textit{i.e.}, flatten and concatenate) the feature maps to a two-dimensional matrix $X$, where each row is a $C_h$-dimensional feature vector of one pixel from the hierarchical features. After that, we compute the $\mathbf{Q}$ (query), $\mathbf{K}$ (key) and $\mathbf{V}$ (value) by linear projections of $X$ as
\begin{equation}\label{eq:QKV}
   \mathbf{Q}=X\mathbf{P}_Q,~~~~\mathbf{K}=X\mathbf{P}_K,~~~~\mathbf{V}=X\mathbf{P}_V,
\end{equation}
where $\mathbf{P}_Q$, $\mathbf{P}_K$, and $\mathbf{P}_V$ are linear projections, respectively. We attempt to apply the self-attention module to enhance the features. However, extremely numerous feature vectors lead to an unbearable memory cost. To alleviate this issue, we propose to adopt a deformable version that only attends to a limited set of key sampling vectors in a learnable manner. It is reasonable for the depth estimation task since several key points that indicate the scene structure are enough for feature enhancement. Let $q$ and $v$ index a element with representation feature $x_q\in \mathbf{Q}$ and $x_v\in \mathbf{V}$, respectively. $p_q$ reprents the location of the query vector $x_q$. The processing can be formulated as
\begin{equation}\label{eq:deformattn}
   \textrm{DAttn}(x_q, x_v, p_q) =
       \sum\limits_{k\in \Omega_k} A_{qk} \cdot x_v \left(p_q + \Delta p_{qk}\right),
\end{equation}
where the attention weight $A_{qk}$ and the sampling offset $\Delta p_{qk}$ of the $k^{th}$ sampling point are obtained via linear projection over the query feature $x_q$. $A_{qk}$ are normalized as $\sum_{k\in \Omega_k} A_{qk} = 1$. As $p_q + \Delta p_{qk}$ is fractional, bilinear interpolation is applied as in~\citep{dai2017deformable} in computing $x_v(p_q + \Delta p_{qk})$. We also add a hierarchical embedding to identify which feature level each query pixel lies in. The output denoted as $\hat{X}$ is folded (\textit{i.e.}, split and reshaped) back to the original resolutions to get the hierarchical enhanced features ${\mathbf{F}}_{enh}$. After fusing ${\mathbf{F}}_{enh}$ and $\textbf{F}$ via channel-wise concatenatetions followed by 1$\times$1 convolutions, we obtain the output $\textbf{F}_o=\{{f_o^n}\}_{n=1}^N$ and achieve the feature enhancement.

When the additional convolution branch is available, we consider a feature map $\mathbf{G}$ as the second input of the HAHI for affinity modeling. Similar to the first input $\mathbf{F}$, $\mathbf{G}$ can be any other type of representation. We utilize a 1$\times$1 convolution to project $\mathbf{G}$ to $\mathbf{G}_h$ with a channel dimension $C_h$ and then flatten $\mathbf{G}_h$ to a two-dimensional query matrix $\mathbf{Q}$. Applying $\hat{X}$ as $\mathbf{K}$ and $\mathbf{V}$, we calculate the cross-attention to model the affinity. Similarly, the unbearable memory cost still persists. We apply the deformable attention module in Eq.~\ref{eq:deformattn} to alleviate this issue, where the reference point locations $p_q$ are dynamically predicted from the affinity query embedding via a learnable linear projection followed by a sigmoid function. After reshaping the result to the original resolution to form the attentive representation $\mathbf{G}_{att}$, we fuse $\mathbf{G}_{att}$ and $\mathbf{G}$ by a channel-wise concatenation and a 1$\times$1 convolution, getting another output of HAHI, denoted as $\mathbf{G}_o$. This process achieves the affinity modeling and the feature interaction between the Transformer and the CNN branches.

All the outputs of the HAHI (\textit{i.e.}, $\textbf{F}_o$ and $\mathbf{G}_o$) are sent to the baseline decoder~\citep{alhashim2018densedepth, li2021simipu} for depth estimation, which consists of several consecutive UpConv layers that are illustrated in the right part of Fig.~\ref{fig:arch}. The network optimization loss updated from~\citep{eigen2014depth} is:
\begin{equation}
   \mathcal{L}_{pixel} = \alpha \sqrt{\frac{1}{T}\sum_i h_{i}^{2} - \frac{\lambda}{T^2}(\sum_i h_i)^2},
   \label{eq:loss}
\end{equation}
where $h_i = \log \Tilde{d_i} - \log d_i$ with the ground truth depth $d_i$ and predicted depth $\Tilde{d_i}$. $T$ denotes the number of pixels having valid ground truth values. Following~\citep{bhat2021adabins}, we use $\lambda = 0.85$ and $\alpha = 10$ for all experiments.

\begin{table*}[t]
    \centering
    \begin{tabularx}{0.9\linewidth}{@{}l*{6}{C}cc@{}}
            \toprule
            Method        & \textbf{$\delta_1$}$\uparrow$       & \textbf{$\delta_2$}$\uparrow$          & \textbf{$\delta_3$}~$\uparrow$            & REL~$\downarrow$          & RMS~$\downarrow$  & $log_{10}$~$\downarrow$ & ~~~Reference~~~\\ \midrule
            
            Eigen~\etal~\citep{eigen2014depth}& 0.769 & 0.950 & 0.988 & 0.158 & 0.641 & -- & NIPS2014\\
            Laina~\etal~\citep{laina2016deeper}& 0.811 & 0.953 & 0.988 & 0.127 & 0.573 & 0.055 & 3DV2016 \\
            StructDepth~\citep{li2021structdepth} & 0.817 & 0.955 & 0.988 & 0.140 & 0.534 & 0.060 & ICCV2021 \\
            MonoIndoor~\citep{ji2021monoindoor} & 0.823 & 0.958 & 0.989 & 0.134 & 0.526 & - & ICCV2021 \\
            DORN~\etal~\citep{fu2018deep}& 0.828 & 0.965 & 0.992 & 0.115 & 0.509 & 0.051 & CVPR2018 \\
            BTS~\citep{lee2019bts}& 0.885 & 0.978 & 0.994 & 0.110 & 0.392 & 0.047 & Arxiv2019\\
            DAV~\citep{huynh2020guiding}& 0.882 & 0.980 & 0.996 & 0.108 & 0.412 & -- & ECCV2020\\
            TransDepth~\citep{yang2021transdepth}& 0.900 & 0.983 & 0.996 & 0.106 & 0.365 & 0.045 & ICCV2021\\
            DPT~\citep{ranftl2021dpt} & \underline{0.904} & \underline{0.988} & \textbf{0.998} & 0.110 & \underline{0.357} & 0.045 & ICCV2021 \\
            AdaBins~\citep{bhat2021adabins} & 0.903 & 0.984 & \underline{0.997} & \underline{0.103} & 0.364 & \underline{0.044} & CVPR2020 \\
            \midrule
            \textbf{DepthFormer} & \textbf{0.921} & \textbf{0.989} & \textbf{0.998} & \textbf{0.096} & \textbf{0.339} & \textbf{0.041} & -\\
            \bottomrule
        \end{tabularx}
    \caption{Comparison of performances on the NYU-Depth-v2 dataset. The reported numbers are from the corresponding original papers.}
    \label{tab:results-nyu}
\end{table*}
        
\begin{table*}[t]
    \centering
        \begin{tabularx}{\textwidth}{@{}ll*{6}{C}c@{}}
            \toprule
            Method & Backbone & \textbf{$\delta_1$}$\uparrow$ & \textbf{$\delta_2$}$\uparrow$ & \textbf{$\delta_3$}$\uparrow$ & REL~$\downarrow$ & Sq~$\downarrow$ & RMS~$\downarrow$ & RMS log~$\downarrow$\\
            \midrule
            Godard~\etal~\citep{godard2017unsupervised}  & ResNet-50 & 0.861 & 0.949 & 0.976 & 0.114   & 0.898  & 4.935 & 0.206\\
            Johnston~\etal~\citep{johnston2020self} & ResNet-101 & 0.889 & 0.962 & 0.982 & 0.106 & 0.861 & 4.699 & 0.185\\
            Gan~\etal~\citep{gan2018monocular} & ResNet-101 & 0.890 & 0.964 & 0.985 & 0.098   & 0.666  & 3.933 & 0.173\\
            DORN~\etal~\citep{fu2018deep}  & ResNet-101 & 0.932 & 0.984 & 0.994 & 0.072   & 0.307  & 2.727 & 0.120\\
            Yin~\etal~\citep{yin2019enforcing}  & ResNext-101 & 0.938 & 0.990  & \underline{0.998} & 0.072 & -- & 3.258 & 0.117\\
            PGA-Net~\citep{xu2020probabilistic} & ResNet-50 & 0.952 & 0.992 & \underline{0.998} & 0.063 & 0.267 & 2.634 & 0.101\\
            BTS~\citep{lee2019bts}  &DenseNet-161& 0.956 & 0.993 & \underline{0.998} & 0.059   & 0.245  & 2.756 & 0.096\\
            TransDepth~\citep{yang2021transdepth}& ViT-B+ResNet-50& 0.956 & 0.994 & \textbf{0.999} & 0.064 & 0.252  & 2.755 & 0.098\\
            DPT-Hybrid~\citep{ranftl2021dpt} &ViT-B+R-50-C$_1$C$_2$& 0.959 & \underline{0.995} & \textbf{0.999} & 0.062 & --& 2.573 & 0.092\\
            AdaBins~\citep{bhat2021adabins} &Mini-ViT+E-B5& \underline{0.964} & \underline{0.995} & \textbf{0.999} & \underline{0.058}  & \underline{0.190}  & \underline{2.360} & \underline{0.088}\\
            \midrule
            \textbf{DepthFormer} & Swin-L+R-50-C$_1$ & \textbf{0.975} & \textbf{0.997} & \textbf{0.999} & \textbf{0.052} & \textbf{0.158} & \textbf{2.143} & \textbf{0.079} \\ 
            \bottomrule
        \end{tabularx} 
    \vspace{-0.1cm}
    \caption{Comparison of performances on the KITTI validation dataset. The reported numbers are from the corresponding original papers. Measurements are made for the depth range from $0m$ to $80m$. Best / Second best results are marked \textbf{bold} / \underline{underlined}. R-50 and E-B5 are short for ResNet-50 and EfficientNet-B5~\citep{tan2019efficientnet}, respectively. C$_i$ represents the $i^{th}$ block of the ResNet-50 network.}
    \vspace{-0.1cm}
    \label{tab:results-kitti-val}
\end{table*}

\begin{figure*}[t]
    \centering
    \footnotesize
    \begin{tabular}{ccc}
        \\
        \includegraphics[width=0.30\linewidth]{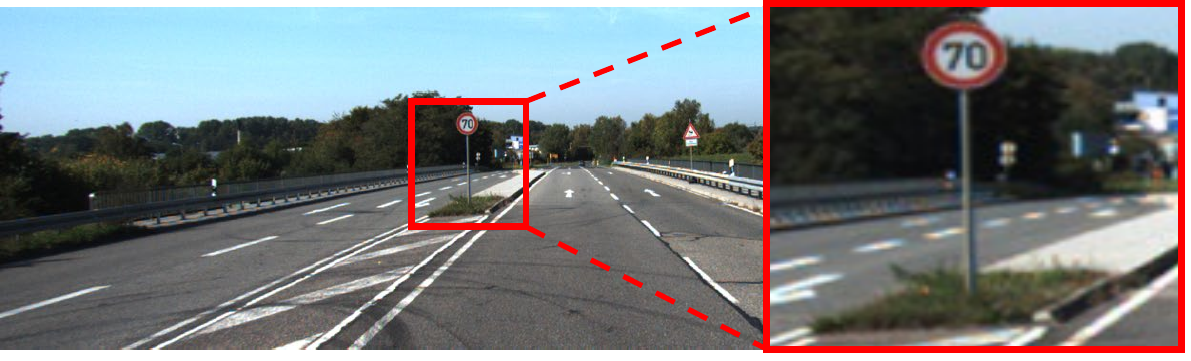}&
        \includegraphics[width=0.30\linewidth]{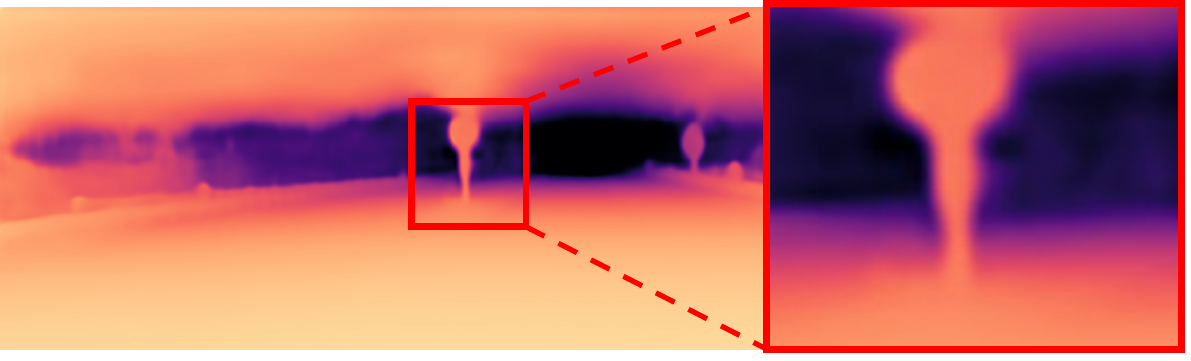}
        \\
        RGB&ResNet-50
        \\
        \includegraphics[width=0.30\linewidth]{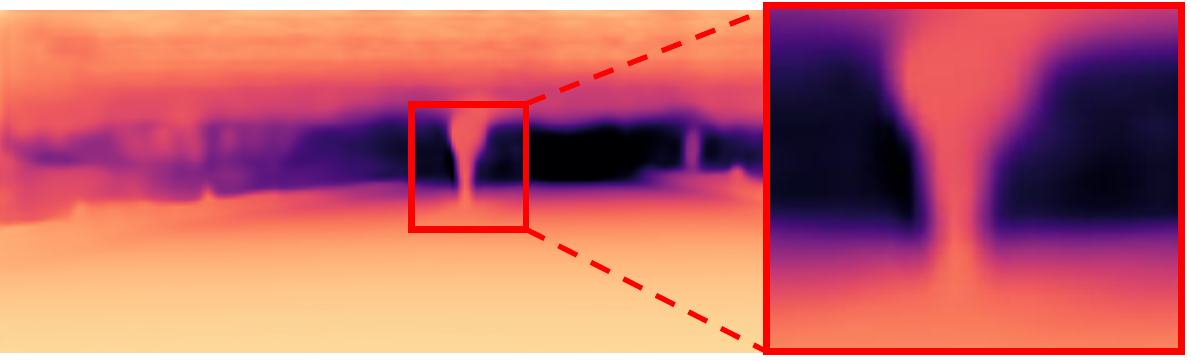}&
        \includegraphics[width=0.30\linewidth]{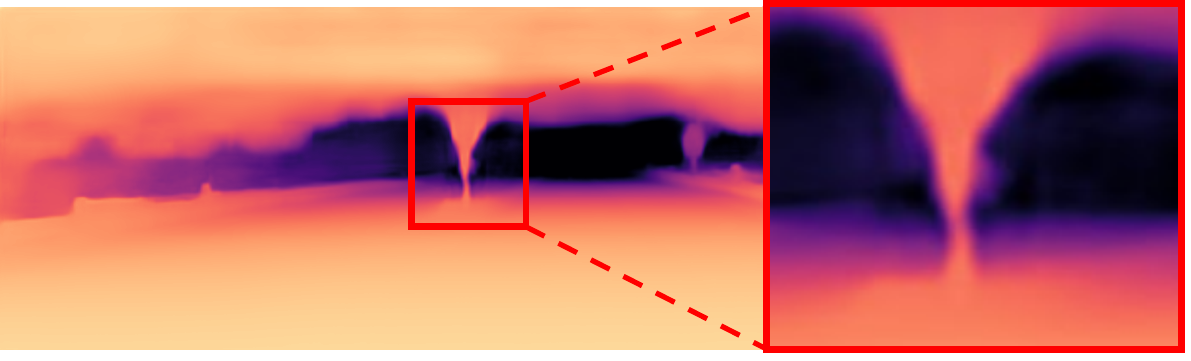}&
        \includegraphics[width=0.30\linewidth]{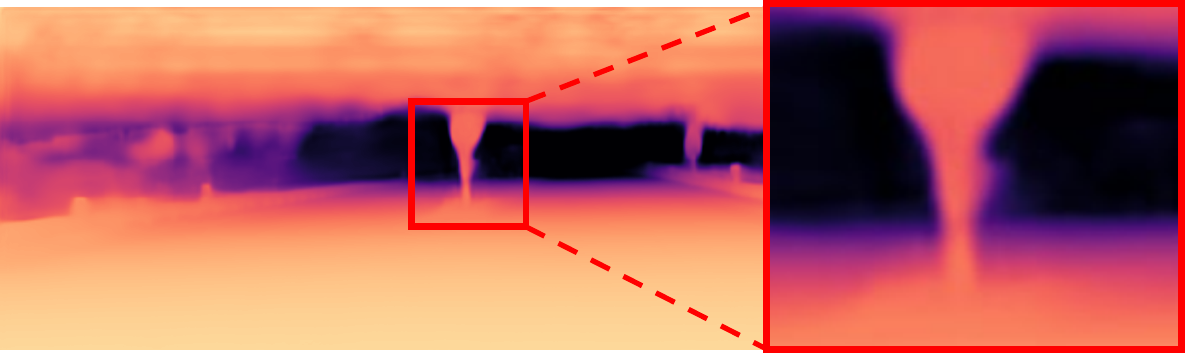}
        \\
        ViT-B&ViT-B+CB&ViT-B+CB+HAHI
        \\
        \includegraphics[width=0.30\linewidth]{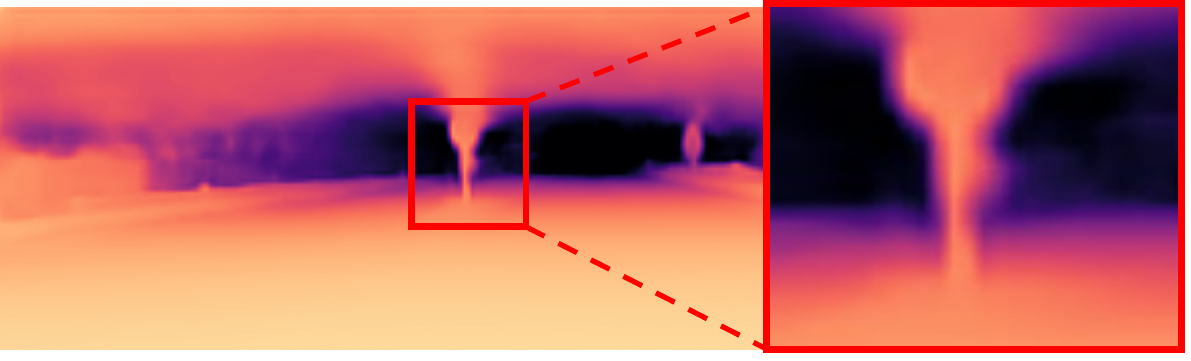}&
        \includegraphics[width=0.30\linewidth]{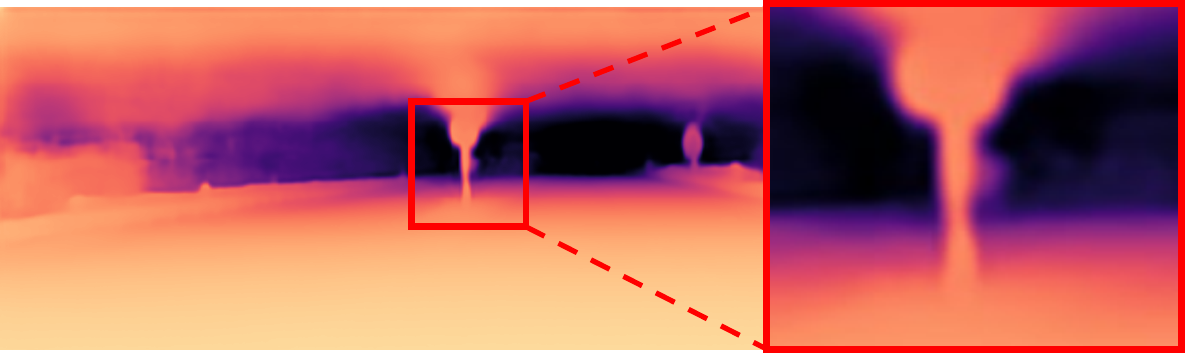}&
        \includegraphics[width=0.30\linewidth]{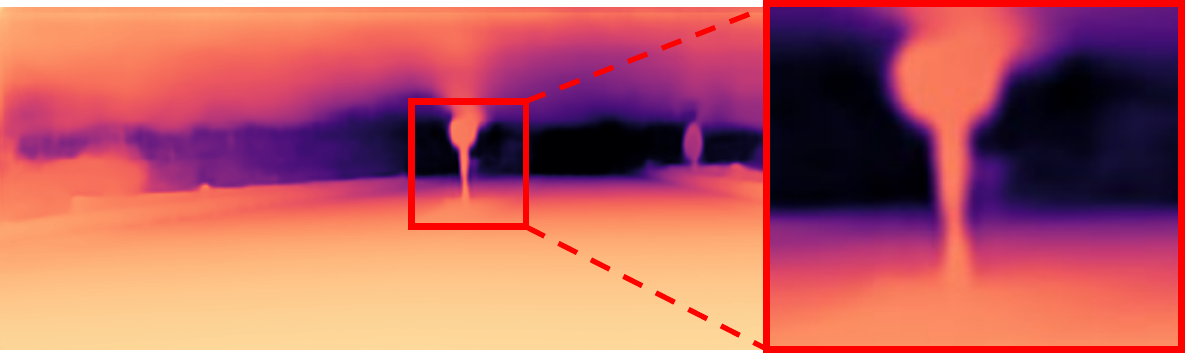}
        \\
        Swin-T&Swin-T+CB&Swin-T+CB+HAHI
        \\
        \includegraphics[width=0.30\linewidth]{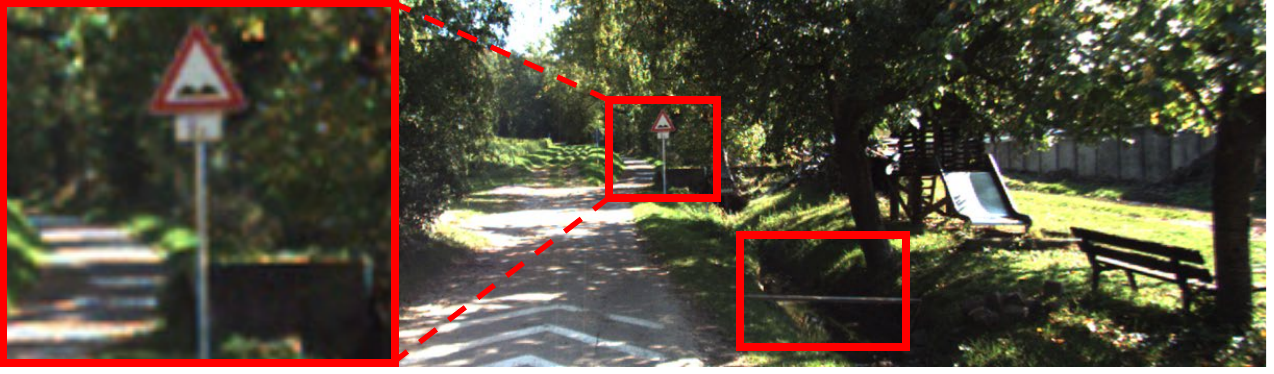}&
        \includegraphics[width=0.30\linewidth]{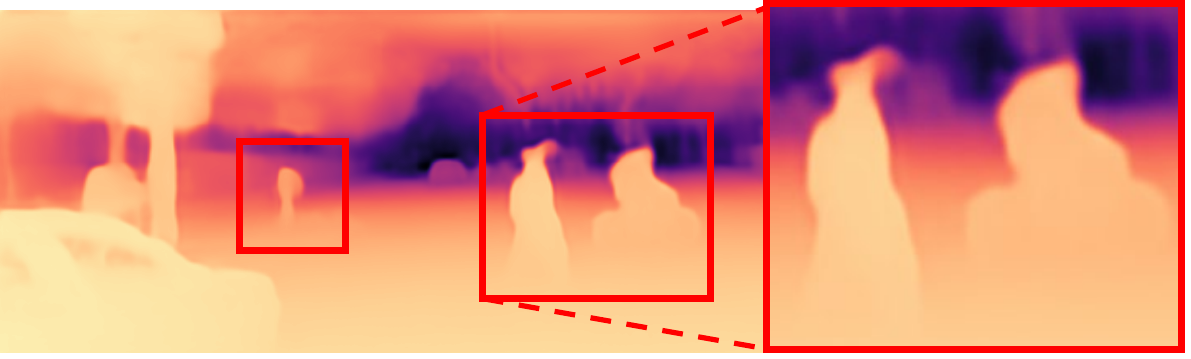}
        \\
        RGB&ResNet-50
        \\
        \includegraphics[width=0.30\linewidth]{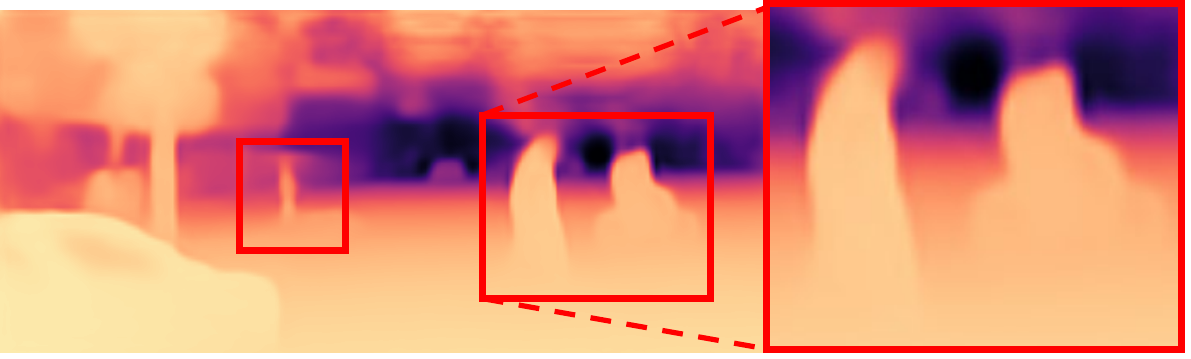}&
        \includegraphics[width=0.30\linewidth]{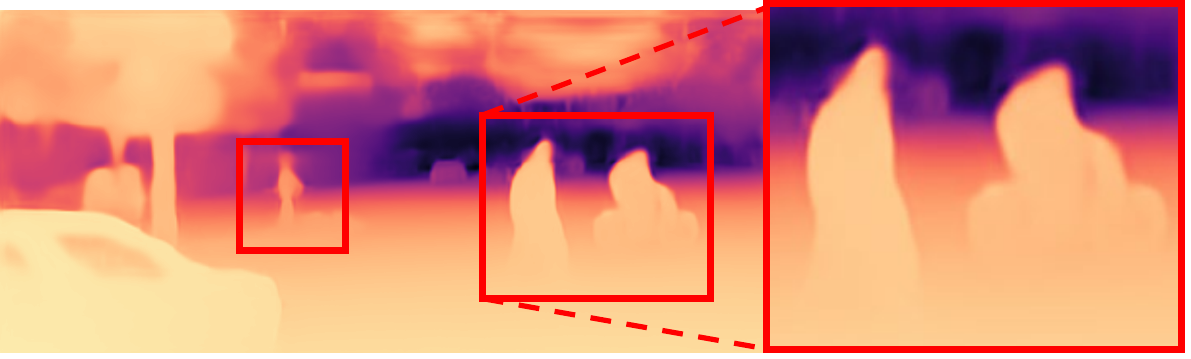}&
        \includegraphics[width=0.30\linewidth]{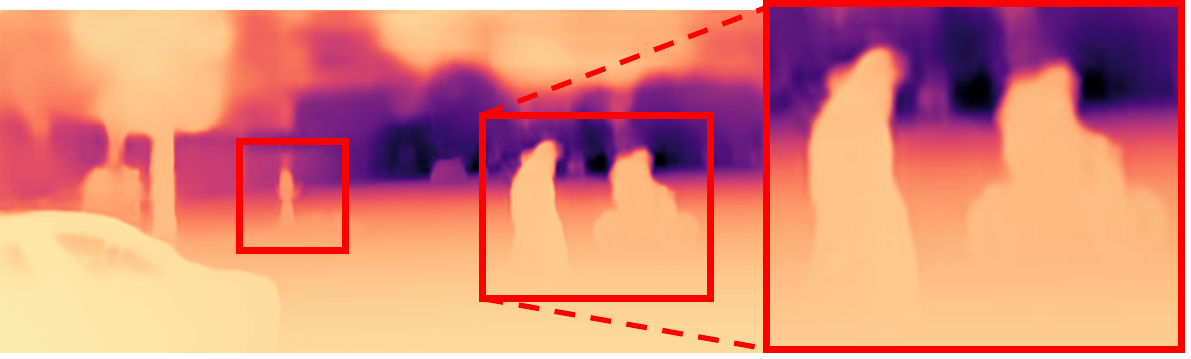}
        \\
        ViT-B&ViT-B+CB&ViT-B+CB+HAHI
        \\
        \includegraphics[width=0.30\linewidth]{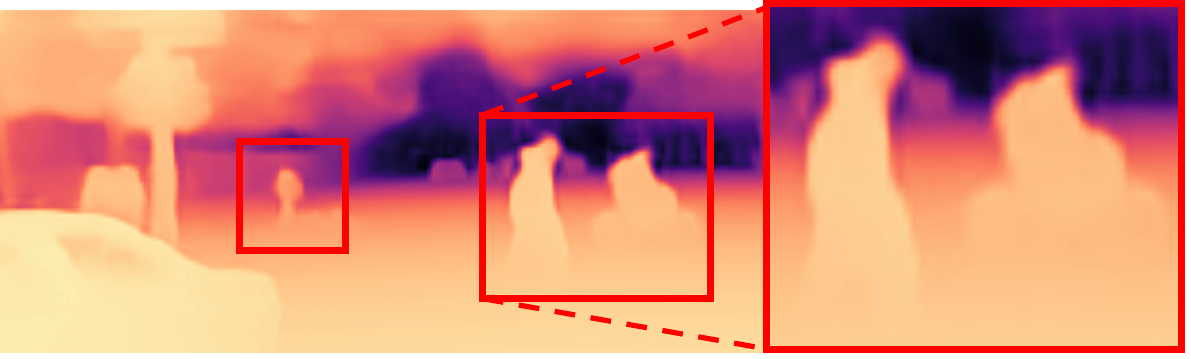}&
        \includegraphics[width=0.30\linewidth]{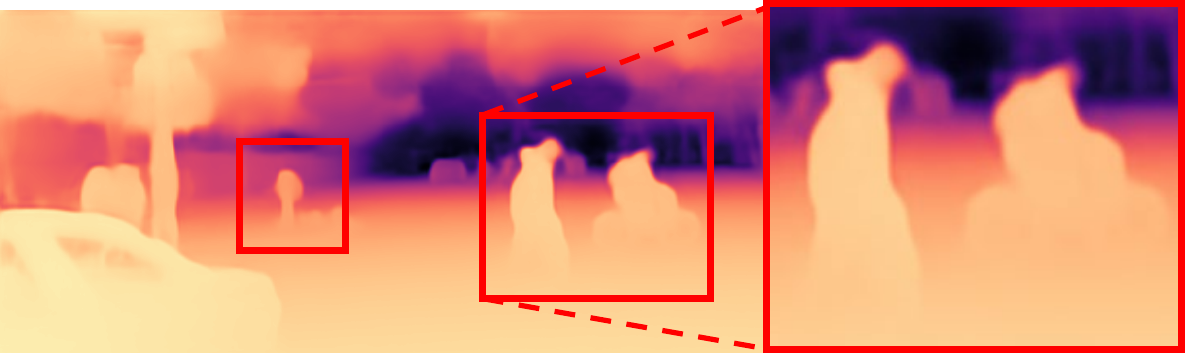}&
        \includegraphics[width=0.30\linewidth]{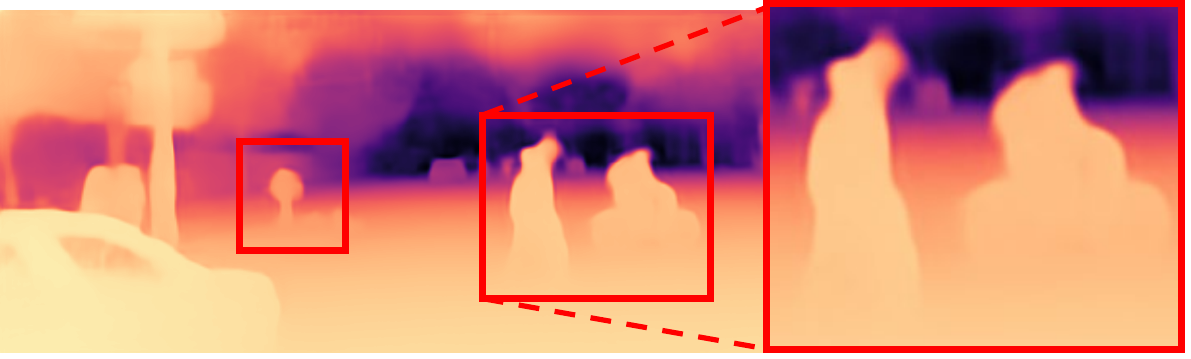}
        \\
        Swin-T&Swin-T+CB&Swin-T+CB+HAHI
        \\
        \includegraphics[width=0.30\linewidth]{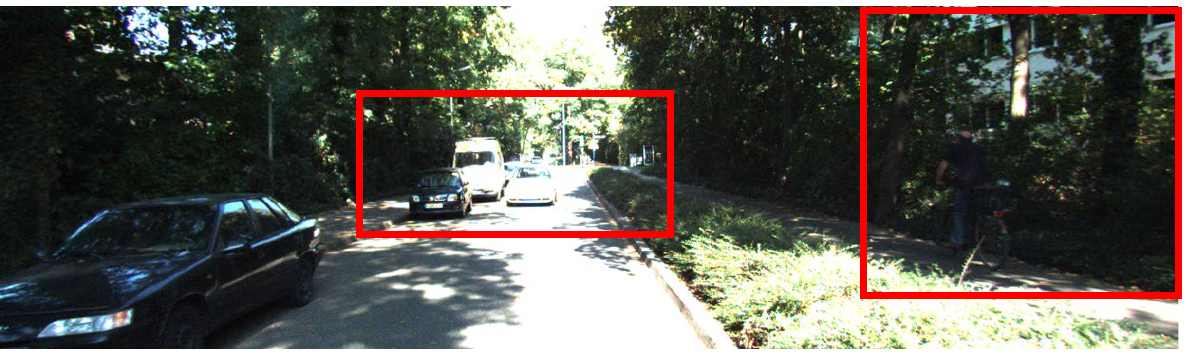}&
        \includegraphics[width=0.30\linewidth]{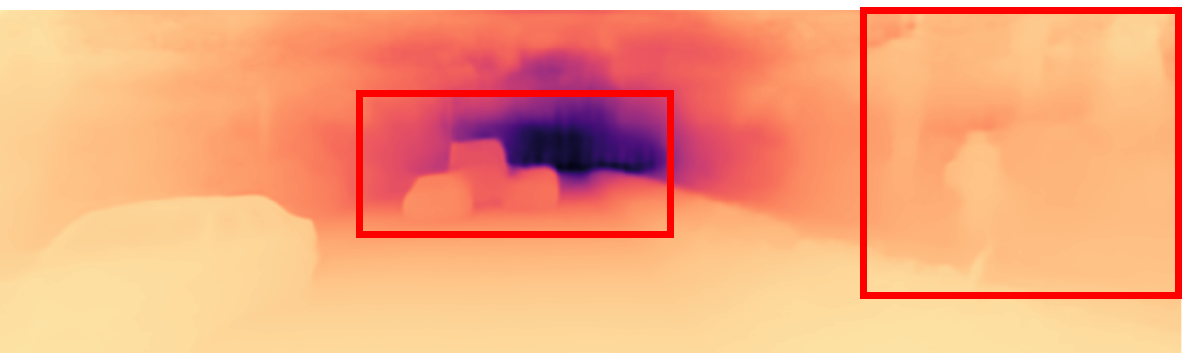}
        \\
        RGB&ResNet-50
        \\
        \includegraphics[width=0.30\linewidth]{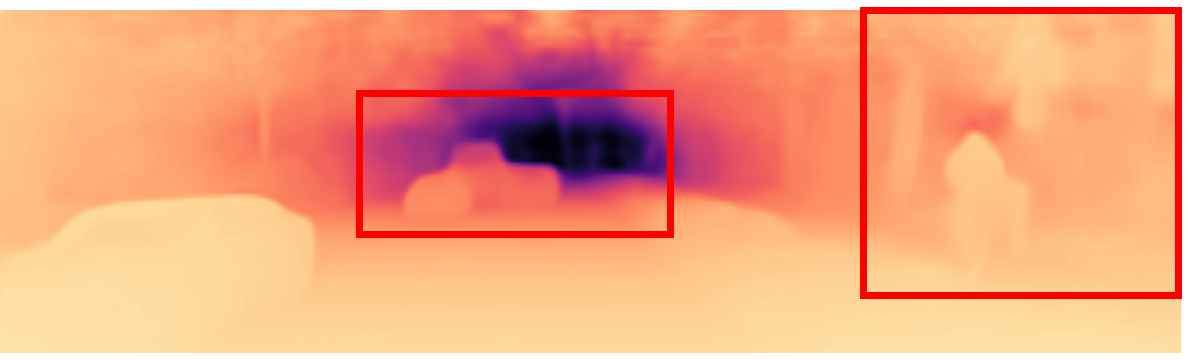}&
        \includegraphics[width=0.30\linewidth]{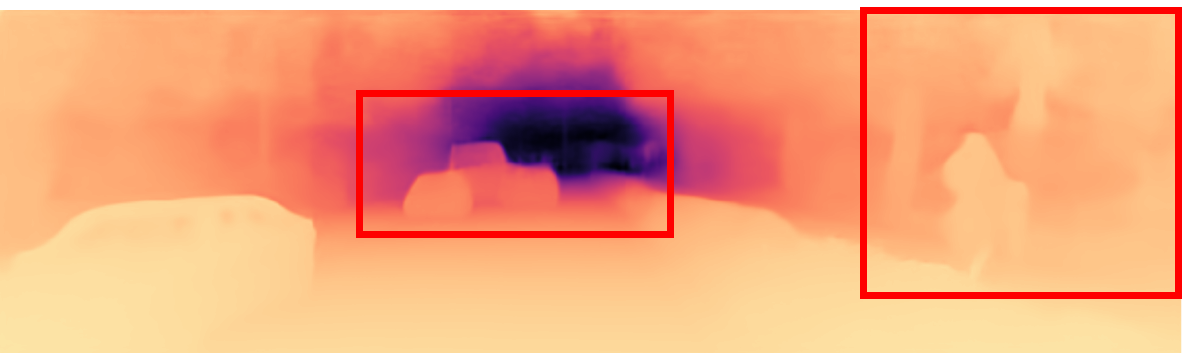}&
        \includegraphics[width=0.30\linewidth]{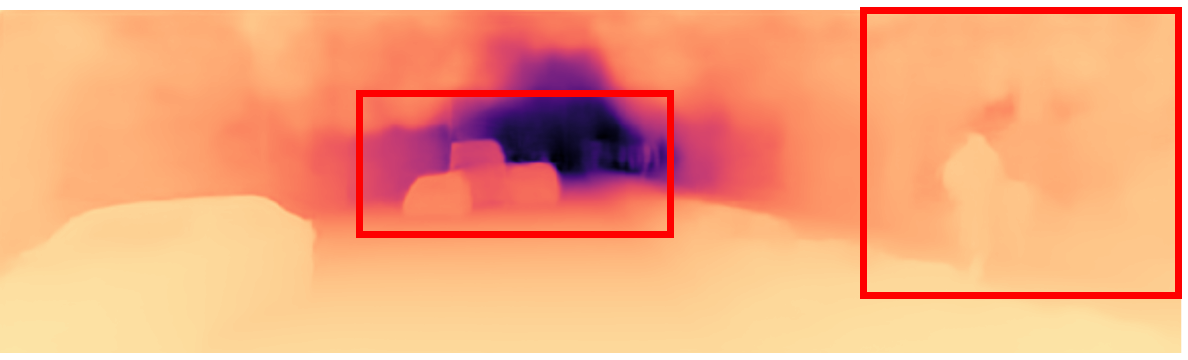}
        \\
        ViT-B&ViT-B+CB&ViT-B+CB+HAHI
        \\
        \includegraphics[width=0.30\linewidth]{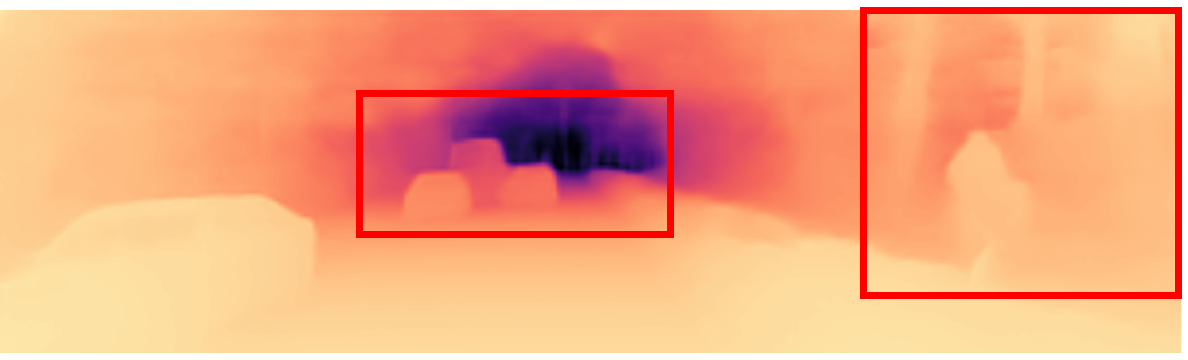}&
        \includegraphics[width=0.30\linewidth]{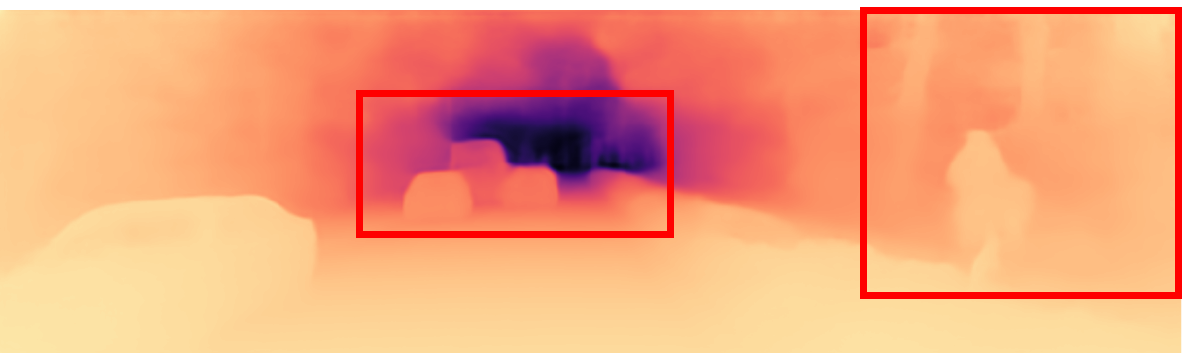}&
        \includegraphics[width=0.30\linewidth]{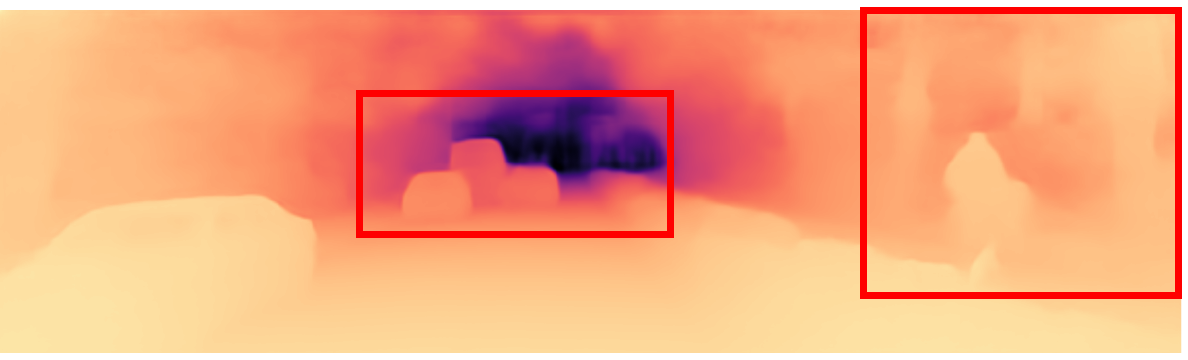}
        \\
        Swin-T&Swin-T+CB&Swin-T+CB+HAHI
        \\
    \end{tabular}
    \caption{Qualitative comparisons in our pilot study.}
    \label{fig:sup-qualitative-comparison-pilot}
 \end{figure*}

\section{Experiment Results}
\label{sec:experiments}
\subsection{Datasets}

\begin{table*}[t]
    \centering
    \begin{adjustbox}{width=0.875\textwidth,center}
        \begin{tabular}{@{}lccccc@{}}
            \toprule
            Method  & ~~~~SILog$\downarrow$~~~~ & sqErrorRel$\downarrow$ & absErrorRel$\downarrow$  & ~~~~iRMSE$\downarrow$~~~~ & ~~~Reference~~~ \\ \midrule
            DORN~\citep{fu2018deep} & 11.77  & 2.23 & 8.78 & 12.98 & CVPR2018\\
            BTS~\citep{lee2019bts}  & 11.67  & 2.21 & 9.04 & 12.23 & Arxiv2019    \\
            BANet~\citep{aich2020BANet} & 11.55  & 2.31 & 9.34 & 12.17  &   Arxiv2020\\
            PWA~\citep{lee2021PWA} & 11.45  & 2.30 & 9.05 & 12.32  &  AAAI2021   \\
            ViP-DeepLab~\citep{qiao2021vip} & \underline{10.80} & \underline{2.19} & \underline{8.94} & \underline{11.77}  & CVPR2021 \\ 
            \midrule
            \textbf{Ours}  & \textbf{10.46}  & \textbf{1.82} & \textbf{8.54} & \textbf{11.17} & - \\
            \bottomrule
        \end{tabular}
    \end{adjustbox}
    \vspace{-0.1cm}
    \caption{Comparison of performances on the KITTI depth estimation benchmark test set. Reported numbers are from the official benchmark website.}
    \vspace{-0.3cm}
\label{tab:results-kitti-test}
\vspace{-0.2cm}
\end{table*}

\textbf{KITTI} is a dataset that provides stereo images and corresponding
3D laser scans of outdoor scenes captured by equipment mounted on a moving vehicle~\citep{geiger2013kitti}. The RGB images have a resolution of around $1241\times376$, while the corresponding ground truth depth maps are of low density. Following the standard Eigen training/testing split~\citep{eigen2014depth}, we use around 26K images from the left view for training and 697 frames for testing. When evaluation, we use the crop as defined by Garg et al.~\citep{garg2016unsupervised} and upsample the prediction to the ground truth resolution. For the online KITTI depth prediction, we use the official benchmark split~\citep{uhrig2017sparsity}, which contains around 72K training data, 6K selected validation data and 500 test data without the ground truth. 


\textbf{NYU-Depth-v2} provides images and depth maps for different indoor scenes captured at a pixel resolution of $640\times480$~\citep{silberman2012nyu}. Following previous works, we train our network on a 50K RGB-Depth pairs subset. The predicted depth maps of DepthFormer have a resolution of $320\times240$ and an upper bound of 10 meters. We upsample them by $2\times$ to match the ground truth resolution during both training and testing. We evaluate the results on the pre-defined center cropping by Eigen et al. \citep{eigen2014depth}. 


\textbf{SUN RGB-D} is an indoor dataset consisting of around 10K images with high scene diversity collected with four different sensors~\citep{song2015sun, xiao2013sun3d, janoch2013category}. We apply this dataset for generalization evaluation. Specifically, we cross-evaluate our NYU pre-trained models on the official test set of 5050 images without further fine-tuning. The depth upper bound is set to 10 meters. Note that this dataset is only for evaluation. We do not train on this dataset.

\begin{figure*}[t!]
    \centering
    \footnotesize
    \begin{tabular}{@{}c@{\hspace{0.1cm}}c@{\hspace{0.1cm}}c@{\hspace{0.1cm}}c@{\hspace{0.1cm}}c@{\hspace{0.1cm}}c@{}}
        \includegraphics[width=0.16\linewidth]{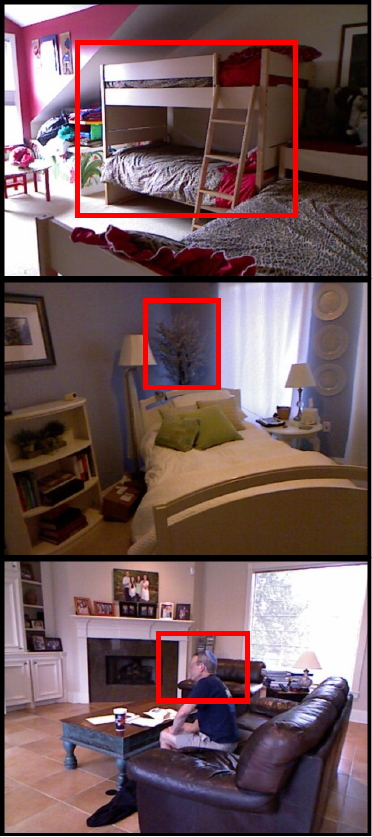} & 
        \includegraphics[width=0.16\linewidth]{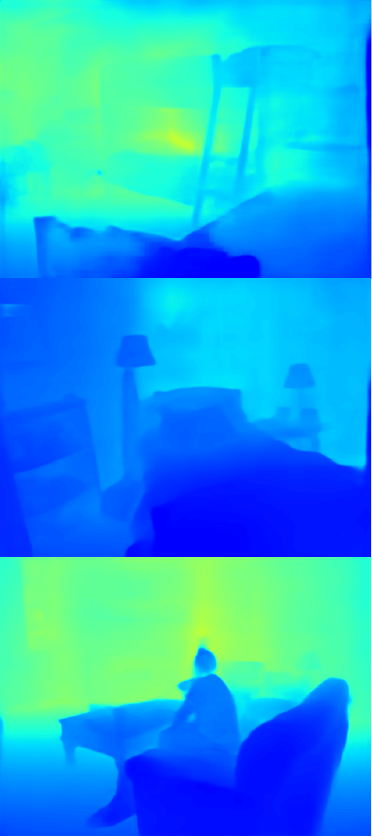} & 
        \includegraphics[width=0.16\linewidth]{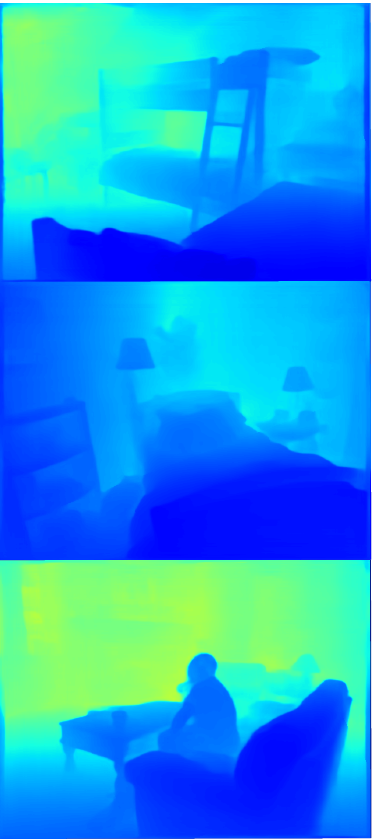} & 
        \includegraphics[width=0.16\linewidth]{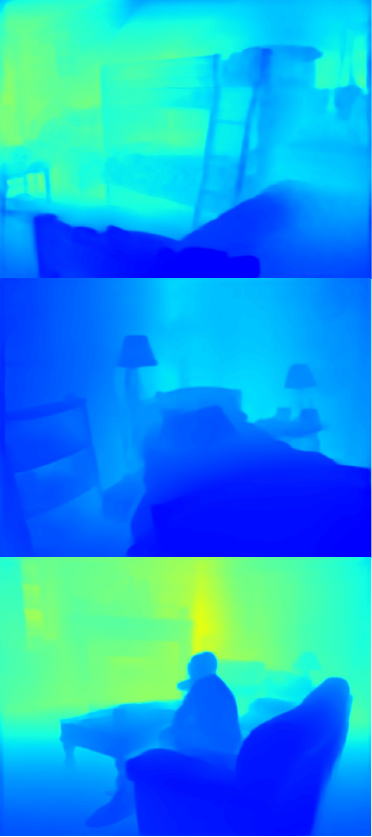} & 
        \includegraphics[width=0.16\linewidth]{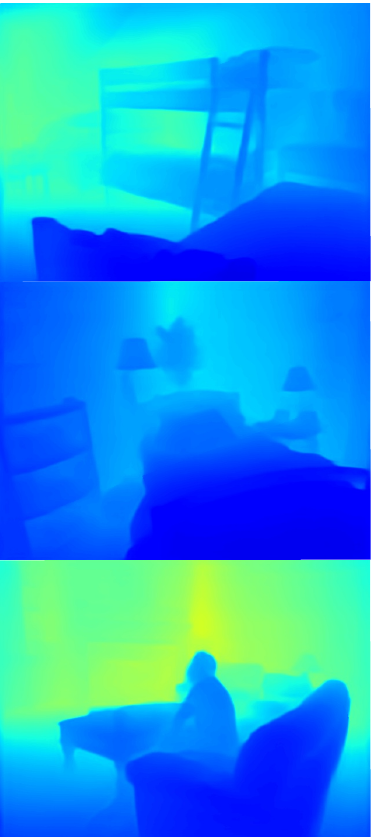} & 
        \includegraphics[width=0.16\linewidth]{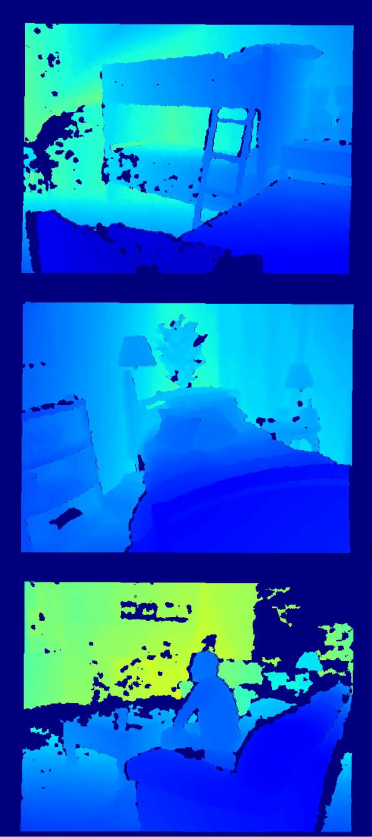} \\
        RGB & BTS~\citep{lee2019bts}& DPT~\citep{ranftl2021dpt} & Ada.~\citep{bhat2021adabins}& Ours& GT\\
    \end{tabular}
    \caption{Qualitative comparison on the NYU-Depth-v2 dataset.}
    \label{fig:qualitative-comparison-nyu}
\end{figure*}

\begin{figure*}[t!]
    \centering
    \footnotesize
    \begin{tabular}{c}
        \includegraphics[width=0.995\linewidth]{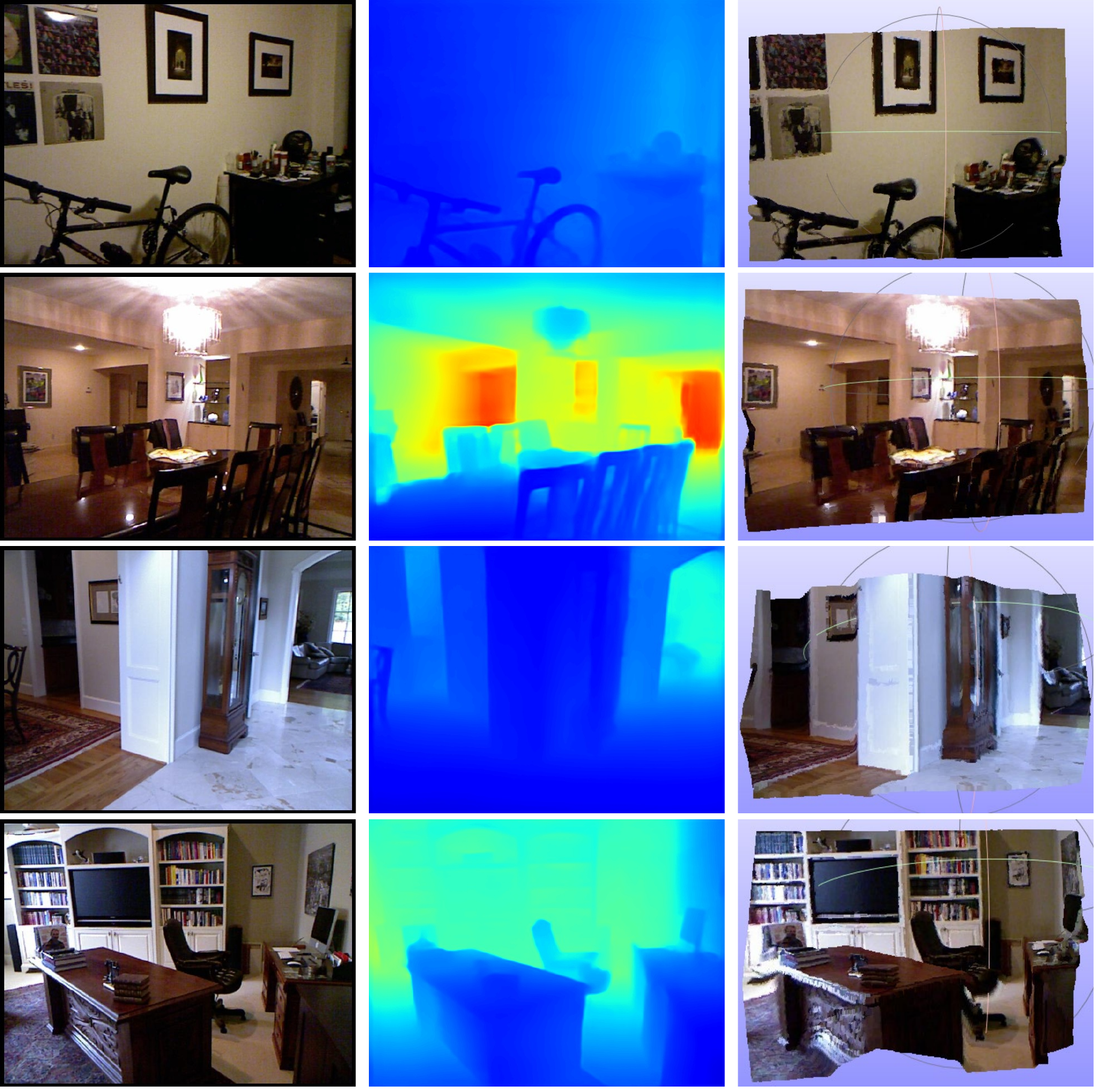} \\
        Input RGB  \hspace{0.2\linewidth} Predicted Depth  \hspace{0.2\linewidth} Reconstructed Point Cloud\\
    \end{tabular}
    \caption{Visualization of reconstructed 3D scenes.}
    \label{fig:pc}
 \end{figure*}

\begin{figure*}[t]
    \centering
    \footnotesize
    \begin{tabular}{@{}c@{\hspace{0.1cm}}c@{\hspace{0.1cm}}c@{\hspace{0.1cm}}c@{}}
        \includegraphics[width=0.33\linewidth]{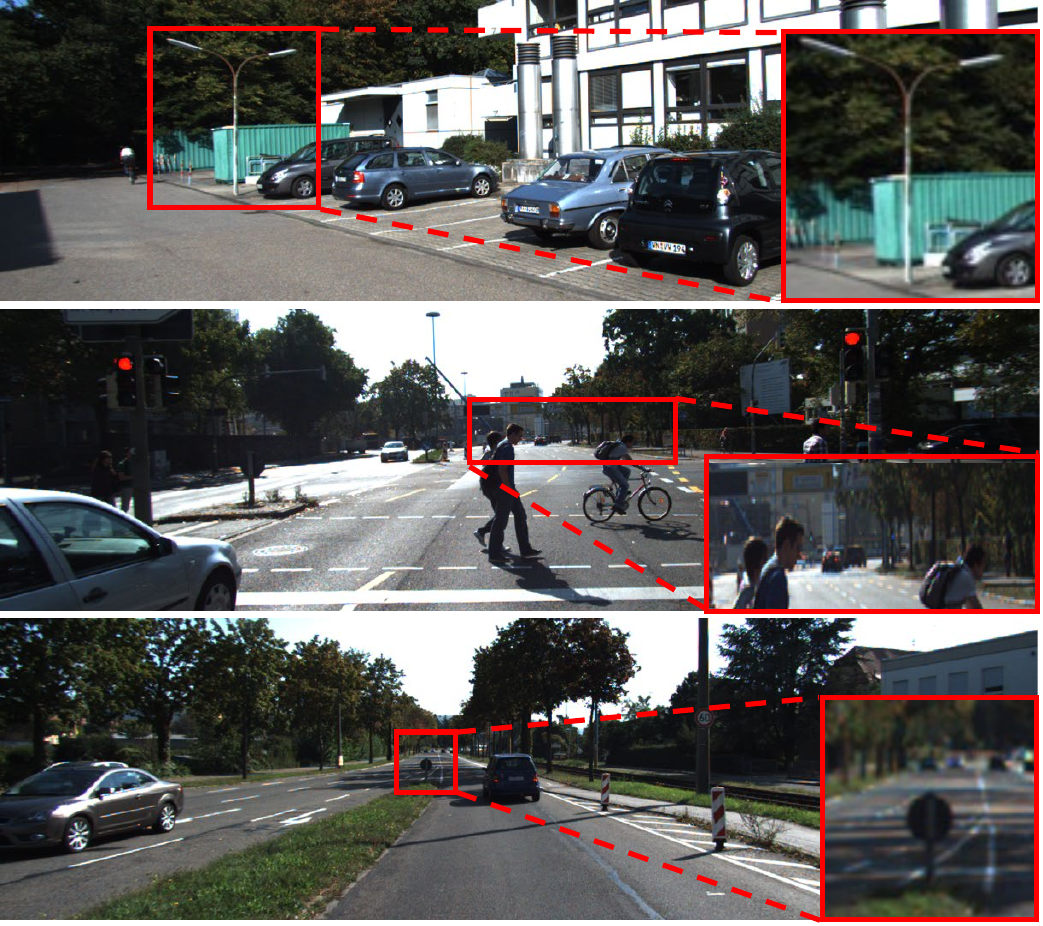} & 
        \includegraphics[width=0.33\linewidth]{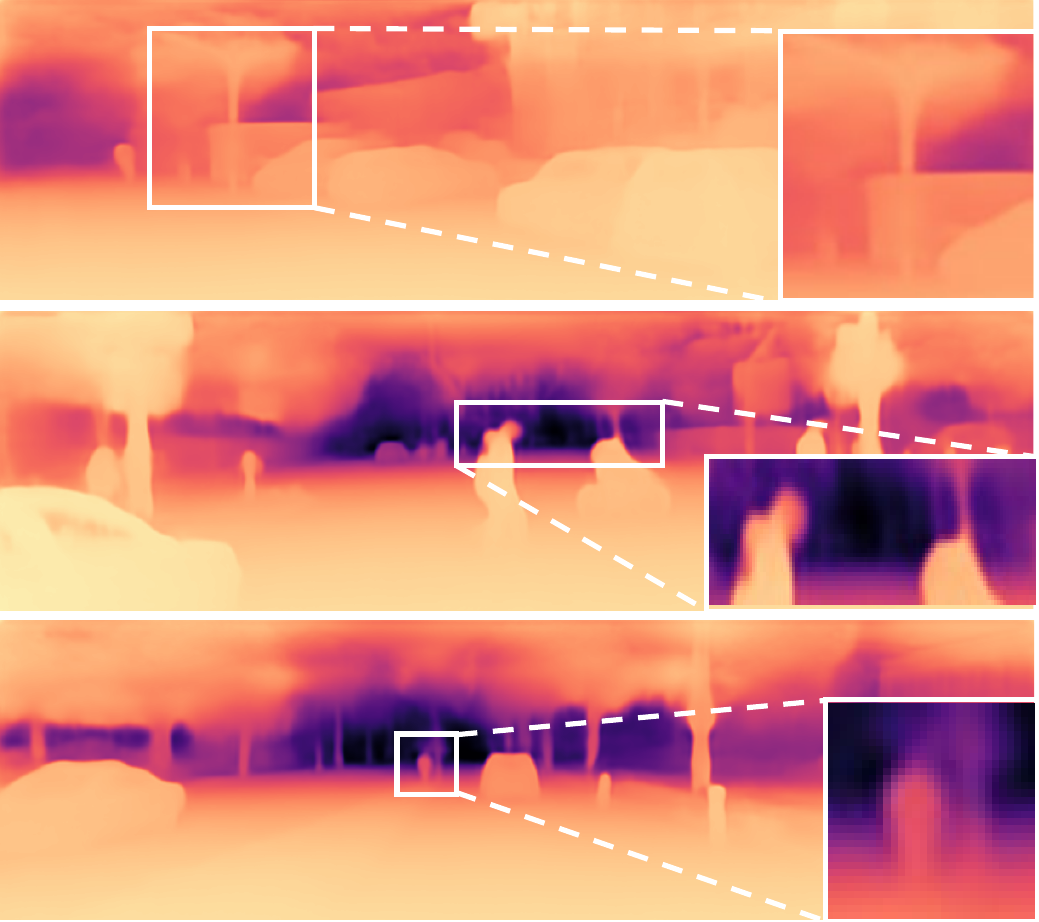} & 
        \includegraphics[width=0.33\linewidth]{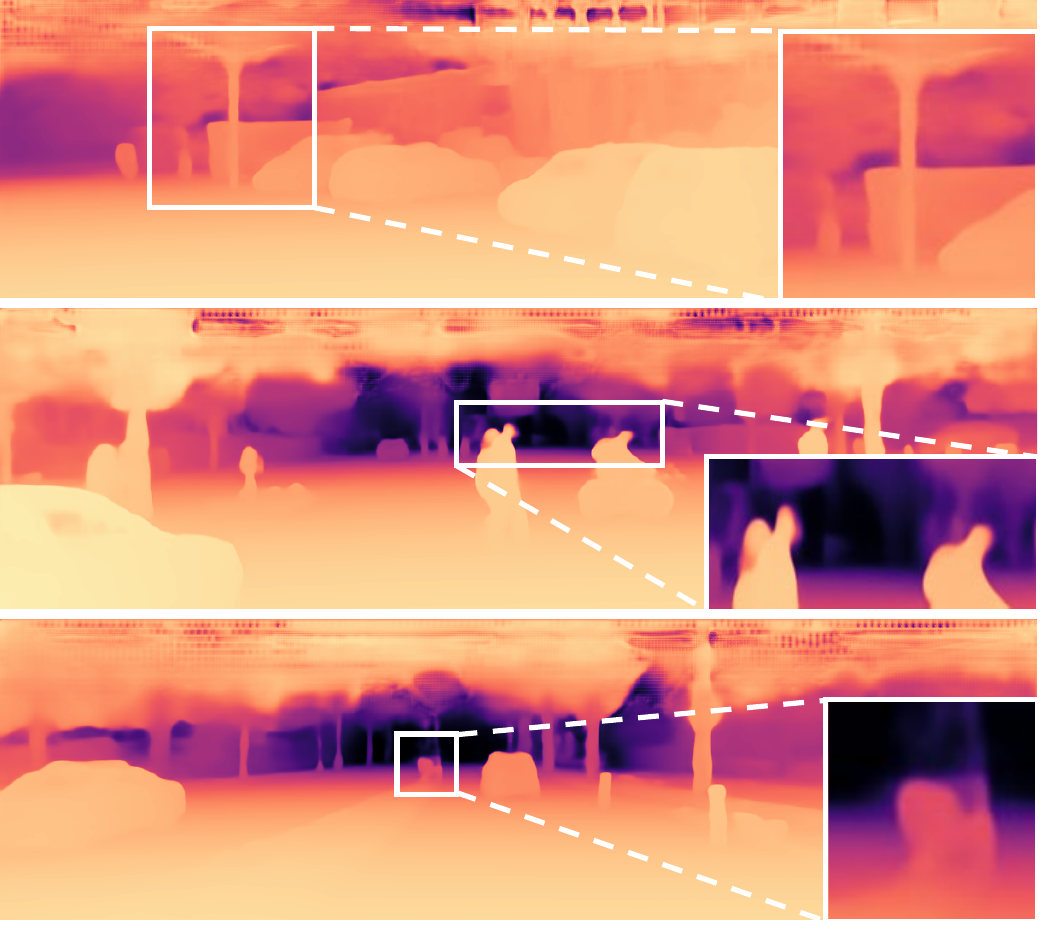} \\ 
        RGB & DenseDepth~\citep{Alhashim2018} & BTS~\citep{lee2019bts}\\
        \includegraphics[width=0.33\linewidth]{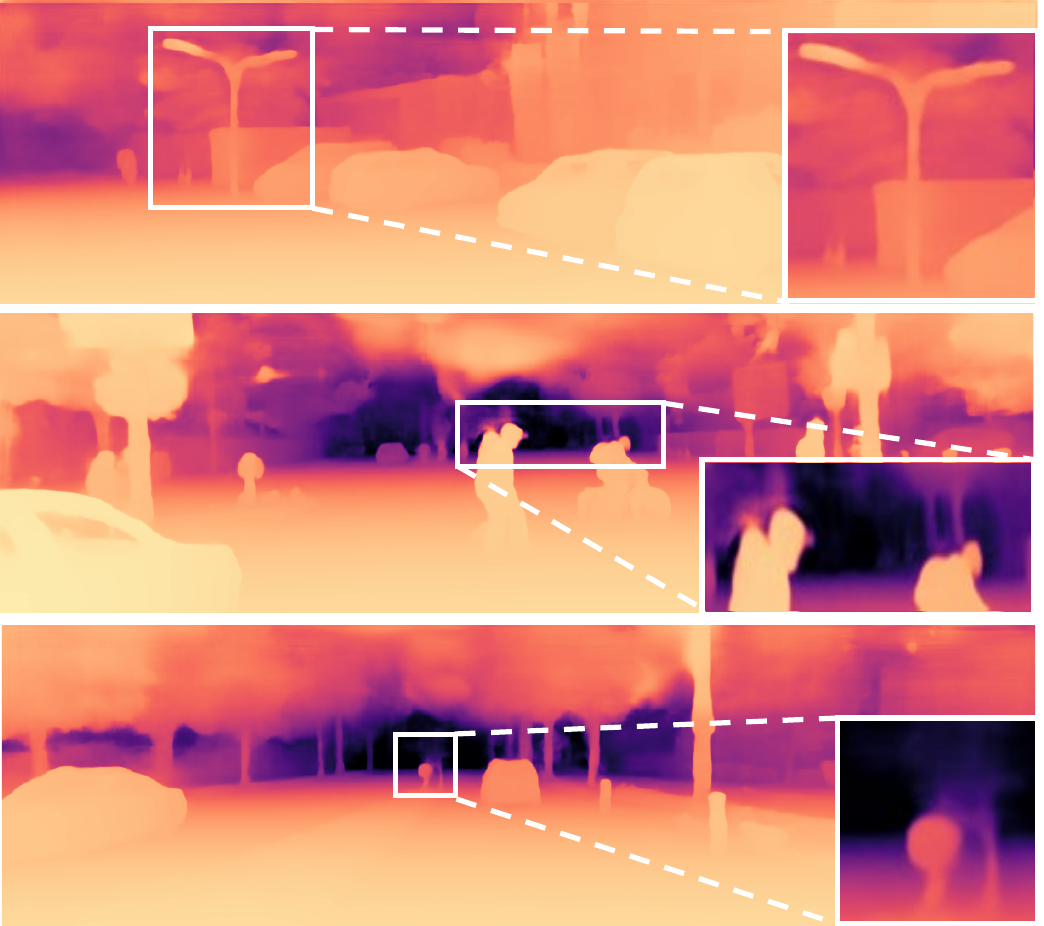} & 
        \includegraphics[width=0.33\linewidth]{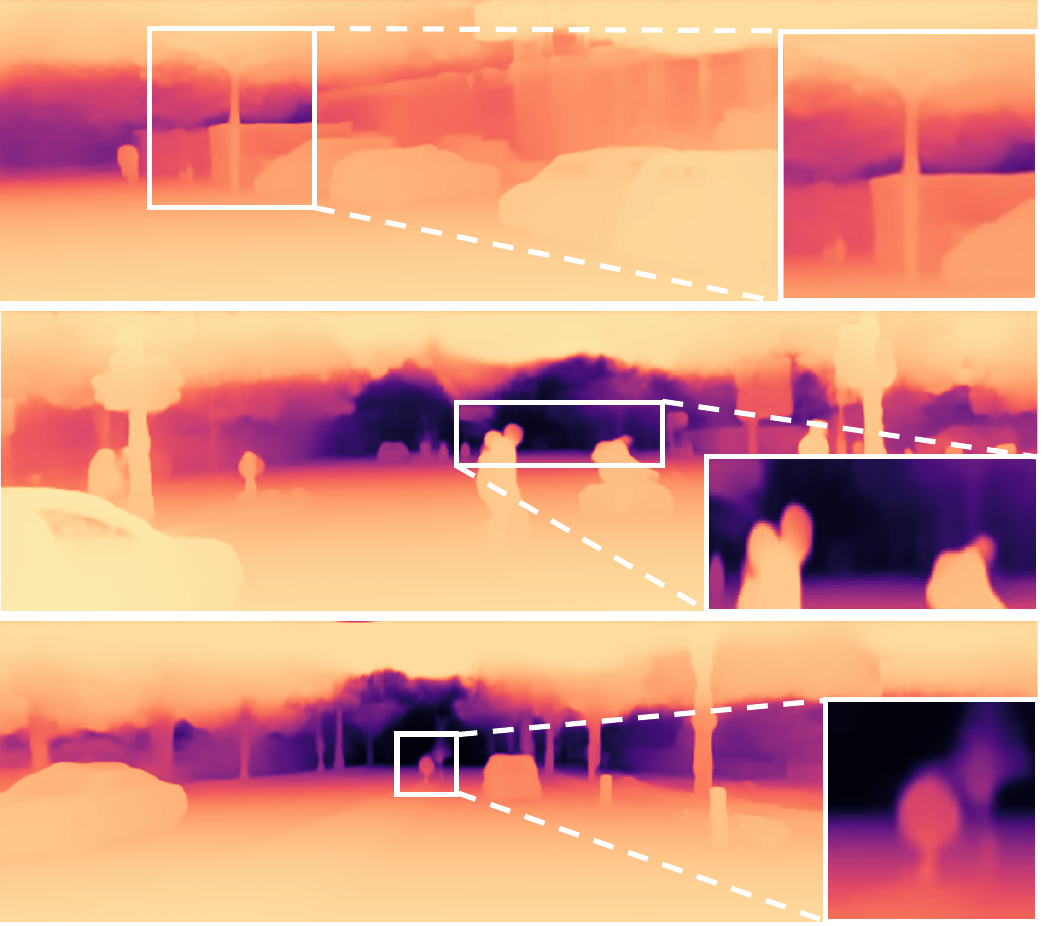} &
        \includegraphics[width=0.33\linewidth]{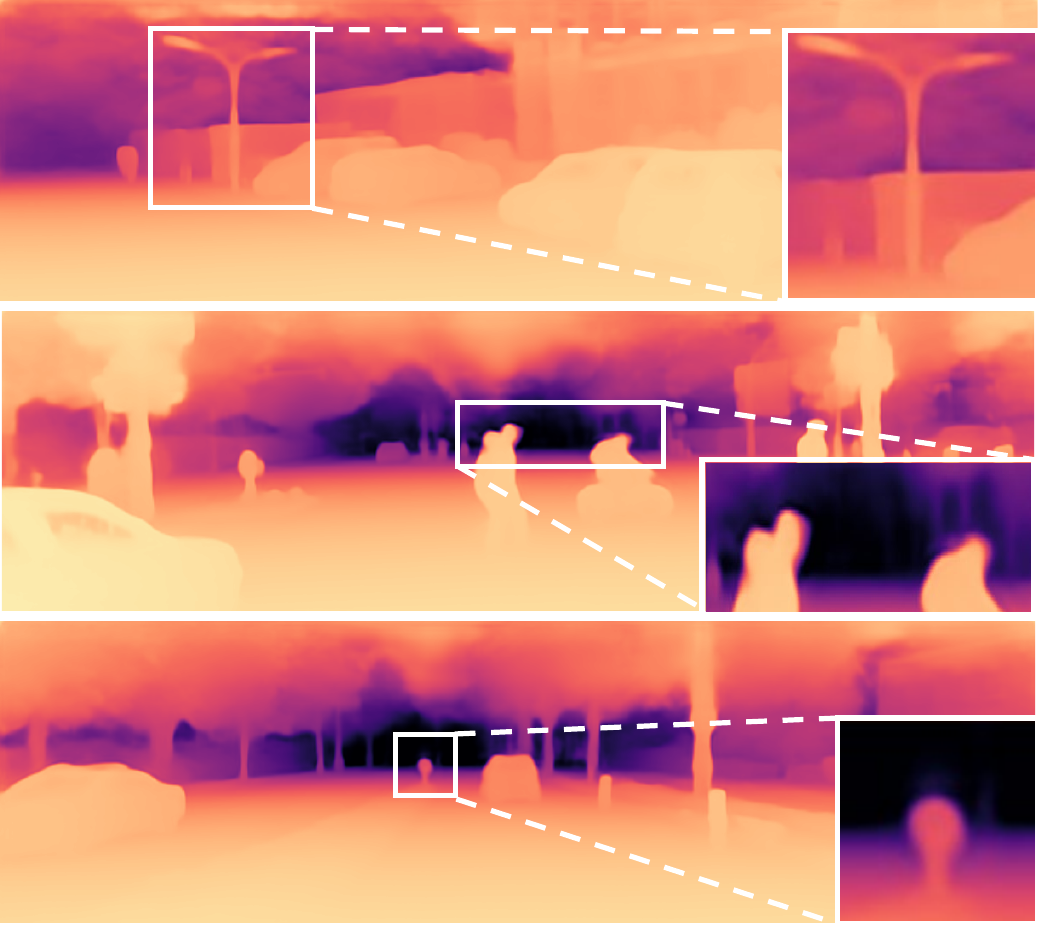}\\ 
        DPT~\citep{ranftl2021dpt} & AdaBins~\citep{bhat2021adabins}& Ours \\
    \end{tabular}
    \caption{Qualitative comparison on the KITTI validation dataset.}
    \label{fig:qualitative-comparison-kitti}
 \end{figure*}

\begin{table*}[htbp]
\centering
    \begin{adjustbox}{width=0.750\linewidth,center}
        \begin{tabular}{@{}lcccccc@{}}
            \toprule
            Method & $\delta_1\uparrow$ & $\delta_2\uparrow$ & $\delta_3\uparrow$ & REL$\downarrow$ & RMS$\downarrow$ & $log_{10}\downarrow$\\ \midrule
            Chen~\etal~\citep{chen2019structure} & 0.757 & 0.943 & \underline{0.984} & 0.166 & 0.494 & 0.071\\
            Yin~\etal~\citep{yin2019enforcing} & 0.696 & 0.912 & 0.973 & 0.183 & 0.541 & 0.082\\
            BTS~\citep{lee2019bts} & 0.740 & 0.933 & 0.980 & 0.172 & 0.515 &  0.075 \\ 
            Adabins~\citep{bhat2021adabins} & \underline{0.771} & \underline{0.944} & 0.983 & \underline{0.159} & \underline{0.476} & \underline{0.068} \\ 
            \midrule
            \textbf{Ours} & \textbf{0.815} & \textbf{0.970} & \textbf{0.993} & \textbf{0.137} & \textbf{0.408} & \textbf{0.059} \\
            \bottomrule
        \end{tabular}
    \end{adjustbox}
    \caption{Results of models trained on the NYU-Depth-v2 dataset and tested on the SUN RGB-D dataset \citep{song2015sun} without fine-tuning. The reported numbers are from~\citep{bhat2021adabins}.}
    \label{tab:generalization}
\end{table*}

\begin{table*}[htbp]
    \centering
    \begin{adjustbox}{width=0.750\linewidth,center}
        \begin{tabular}{@{}llccccc@{}}
            \toprule
            Backbone  & Various & \textbf{$\delta_1$}$\uparrow$ & \textbf{$\delta_2$}$\uparrow$ & \textbf{$\delta_2$}$\uparrow$ & REL$\downarrow$ & RMS$\downarrow$ \\ 
            \midrule
            \multirow{3}{*}{R-50~\citep{he2016resenet}}  
            & -- & 0.811 & 0.961 & 0.991 & 0.145 &  0.482 \\
            & +HAHI & 0.866 & 0.976 & 0.994 & 0.122 &  0.411 \\

            & +HAHI+LP & 0.865 & 0.976 & 0.995 & 0.124 & 0.406 \\
            \midrule
            \multirow{4}{*}{Swin-T~\citep{liu2021swin}} 
            & -- & 0.847 & 0.975 & 0.993 & 0.131 & 0.432 \\
            & +HAHI & 0.866 & 0.977 & 0.995 & 0.125 & 0.409 \\
            & +CB & 0.851 & 0.977 & 0.995 & 0.129 & 0.425 \\
            & +CB+HAHI & 0.871 & 0.979 & 0.995 & 0.120 & 0.399 \\
            \midrule
            \multirow{2}{*}{Swin-B~\citep{liu2021swin}} 
            & +CB+HAHI & 0.892 & 0.984 & 0.997 & 0.109 & 0.373 \\
            & +CB+HAHI+LP & 0.910 & 0.987 & 0.997 & 0.101 & 0.348 \\
            \midrule
            Swin-L~\citep{liu2021swin}
            & +CB+HAHI+LP & \textbf{0.921} & \textbf{0.989} & \textbf{0.998} & \textbf{0.096} & \textbf{0.339} \\
            \bottomrule
        \end{tabular}
    \end{adjustbox}
    \caption{Ablation study results on the NYU dataset. CB: Convolution Branch. LP: Larger-scale pre-training dataset (22K ImageNet) for boosting the model  performance. For fair comparison, we utilize the 22K-ImageNet pre-trained ResNet-50-x3 provided by~\citep{kolesnikov2020big} to get the results of R-50 (+HAHI+LP).}
    \label{tab:nyu_ablation}
\end{table*}


\begin{table*}[t]
	\centering
    \begin{adjustbox}{width=0.70\linewidth,center}
		\begin{tabular}{@{} c c c  c c c@{}}
            \toprule
            \multicolumn{2}{c}{Aggregation (DSA)} & \multirow{2}*{Interaction (DCA)} &  \multirow{2}*{\textbf{$\delta_1$$\uparrow$}} & \multirow{2}*{REL$\downarrow$} & \multirow{2}*{RMS$\downarrow$}\\
            ~~~~~~single-level~~~~~~  & ~~~~~~multi-level~~~~~~  &  &\\
            \midrule
            & & & 0.847 & 0.131 & 0.432 \\
            \checkmark& & & 0.850 & 0.131 & 0.427 \\
            & \checkmark & & 0.867 & 0.124 & 0.408 \\
            & & \checkmark & 0.828 & 0.143 & 0.443 \\
            \checkmark & & \checkmark & 0.831 & 0.144 & 0.449 \\
            &\checkmark & \checkmark & \textbf{0.871} & \textbf{0.120} & \textbf{0.399} \\
            \bottomrule
        \end{tabular}
	\end{adjustbox}
	\caption{Ablation study of the HAHI module on NYU dataset. DSA, DCA: Deformable self-attention and deformable cross-attention.}
	\label{tab:abl_hahi}
\end{table*}

\subsection{Evaluation Metrics}
In our experiments, we follow the standard evaluation protocol of the prior work~\citep{eigen2014depth} to confirm the effectiveness of DepthFormer in experiments. For the NYU, KITTI Eigen split and SUN RGB-D dataset, we utilize the accuracy under the threshold ($\delta_i < 1.25^i, i = 1, 2, 3$), mean absolute relative error (AbsRel), mean squared relative error (SqRel), root mean squared error (RMSE), root mean squared log error (RMSElog), and mean log10 error (log10) to evaluate our methods. In terms of the online KITTI benchmark~\citep{uhrig2017sparsity}, we use the scale-invariant logarithmic error (SILog), percentage of AbsRel and SqRel (absErrorRel, sqErrorRel), and root mean squared error of the inverse depth (iRMSE).

\subsection{Implementation Details}
Since we find there is no commonly used codebase for the monocular depth estimation task, we develop a unified benchmark based on the MMSegmentation~\citep{mmseg2020}. We believe it can further boost the development of this field and achieve fair comparisons. We train the entire network with the batch size 2, learning rate $1e^{-4}$ for 38.4k iterations on a single node with 8 NVIDIA V100 32GB GPUs, which takes around 5 hours. The linear learning rate warm-up strategy is applied for the first 30\% iterations following~\citep{bhat2021adabins}. The cosine annealing learning rate strategy is adopted for the learning rate decay. Following~\citep{ranftl2021dpt, liu2021swin}, we sample $N=4$ results from the transformer features as the output of the transformer branch. The number of reference points in deformable attention modules and $C_h$ is experientially set to 8 and the median value of the channel dimension of $\textbf{F}$, respectively. Following~\citep{zhu2020deformabledetr}, we adopt 8 deformable attention heads. The default patch size of ViT-Base and window size of Swin Transformer are 16 and 12, respectively. Following previous works, our encoders are pre-trained on ImageNet dataset~\citep{krizhevsky2012imagenet} and then fine-tuned on depth datasets. As for the pilot study, the baseline model consists of an encoder and a decoder. We adopt the decoder in~\citep{alhashim2018densedepth} as a default setting and mainly focus on the influence of encoder choices. In terms of the pure convolution encoder, we utilize the standard ResNet-50~\citep{he2016resenet}. For the pure Transformer encoder, we adopt the ViT-B~\citep{dosovitskiy2020vit} following the design of the DPT~\citep{ranftl2021dpt} and Swin-T~\citep{liu2021swin}. Notably, our encoders are pre-trained on the ImageNet classification, which is the standard protocol of supervised monocular depth estimation. During training, we adopt the AdamW optimizer. The weight decay is set to 0.01. We experientially use the 1-cycle policy with the learning rate $lr = 6e^{-5}$ for the Transformer-based model and $lr = 1e^{-4}$ for the ResNet-based model. We also apply a linear warm-up scheduler for the first 500 iterations. The cosine annealing learning rate strategy is adopted for the learning rate decay. When evaluation, we divide the depth range to 0-20$m$, 20-60$m$ and 60-80$m$. The results of 0-20$m$ and 60-80$m$ can indicate the model performance predicting the depth of near and distant objects, respectively. We also present more detailed results in Fig.~\ref{fig:sup-distance}, where the model performance at each tick is shown in a curve. When visualizing the results, we utilize the color map of jet and reversed magma for NYU and KITTI, respectively.

\begin{table*}[t]
    \centering
        \begin{tabularx}{0.985\textwidth}{@{}lll*{5}{C}c@{}}
            \toprule
            Backbone  & Various & Range & \textbf{$\delta_1$}$\uparrow$ & \textbf{$\delta_2$}$\uparrow$ & \textbf{$\delta_2$}$\uparrow$ & REL$\downarrow$ & RMS$\downarrow$ \\ 
            \midrule
            \multirow{3}{*}{ResNet-50~\citep{he2016resenet}} &
            \multicolumn{1}{l}{\multirow{3}{*}{-}}
            & 0$m$-20$m$ & 0.973  & 0.998 & 1 & 0.054 &  0.985  \\
            && 60$m$-80$m$ & 0.600 & 0.900 & 0.972  & 0.188 & 14.30  \\
            && Overall & 0.952 & 0.994 & 0.999 & 0.065 &  2.596 \\
            \midrule
            
            \multirow{3}{*}{ViT-Base~\citep{dosovitskiy2020vit}} &
            \multicolumn{1}{l}{\multirow{3}{*}{-}}
            & 0$m$-20$m$ & 0.955 & 0.995 & 0.999 & 0.071 & 1.275   \\
            && 60$m$-80$m$ & 0.727 & 0.936 & 0.985 & 0.150 & 11.86  \\
            && Overall & 0.938 & 0.992 & 0.999 & 0.080 & 2.695  \\
            \hdashline
            \multirow{3}{*}{ViT-Base~\citep{dosovitskiy2020vit}} &
            \multicolumn{1}{l}{\multirow{3}{*}{+CB}}
            & 0$m$-20$m$ & 0.960 & 0.996 & 0.999 & 0.067 & 1.223   \\
            && 60$m$-80$m$ & 0.725 & 0.950 & 0.984 & 0.147 & 11.69  \\
            && Overall & 0.942 & 0.994 & 0.999 & 0.076 & 2.644  \\
            \hdashline
            \multirow{3}{*}{ViT-Base~\citep{dosovitskiy2020vit}} &
            \multicolumn{1}{l}{\multirow{3}{*}{+CB+HAHI}}
            & 0$m$-20$m$ & 0.964 & 0.995 & 0.999 & 0.064 & 1.172  \\
            && 60$m$-80$m$ & 0.712 & 0.946 & 0.984 & 0.150 & 11.90 \\
            && Overall & 0.948 & 0.993 & 0.999 & 0.073 & 2.596  \\
            \midrule

            \multirow{3}{*}{Swin-T~\citep{liu2021swin}} &
            \multicolumn{1}{l}{\multirow{3}{*}{-}}
            & 0m-20m & 0.972 & 0.998 & 1 &  0.050 &  0.948 \\
            && 60m-80m & 0.729 & 0.941 & 0.984 & 0.150 &  11.85 \\
            && Overall & 0.961 & 0.993 & 0.999 & 0.062 & 2.402  \\
            \hdashline
            \multirow{3}{*}{Swin-T~\citep{liu2021swin}} &
            \multicolumn{1}{l}{\multirow{3}{*}{+CB}}
            & 0$m$-20$m$ & 0.979 & 0.998 & 1 & 0.049 & 0.934   \\
            && 60$m$-80$m$ & 0.726 & 0.945 & 0.984 & 0.149 & 11.76  \\
            && Overall & 0.964 & 0.995 & 0.999 & 0.060 & 2.310  \\
            \hdashline
            \multirow{3}{*}{Swin-T~\citep{liu2021swin}} &
            \multicolumn{1}{l}{\multirow{3}{*}{+CB+HAHI}}
            & 0$m$-20$m$ & 0.981 & 0.998 & 1 & 0.049 & 0.911   \\
            && 60$m$-80$m$ & 0.744 & 0.939 & 0.981 & 0.146 & 11.61  \\
            && Overall & \textbf{0.966} & \textbf{0.996} & \textbf{0.999} & \textbf{0.059} & \textbf{2.261}  \\
            \bottomrule
        \end{tabularx}
    \vspace{-0.1cm}
    \caption{More detailed ablation quantitative results on KITTI dataset.}
    \label{tab:kitti_pilot_extend}
\end{table*}

\begin{figure*}[t]
   \centering
   \footnotesize
   \begin{tabular}{ccc}
       \\
       &
       \includegraphics[width=0.30\linewidth]{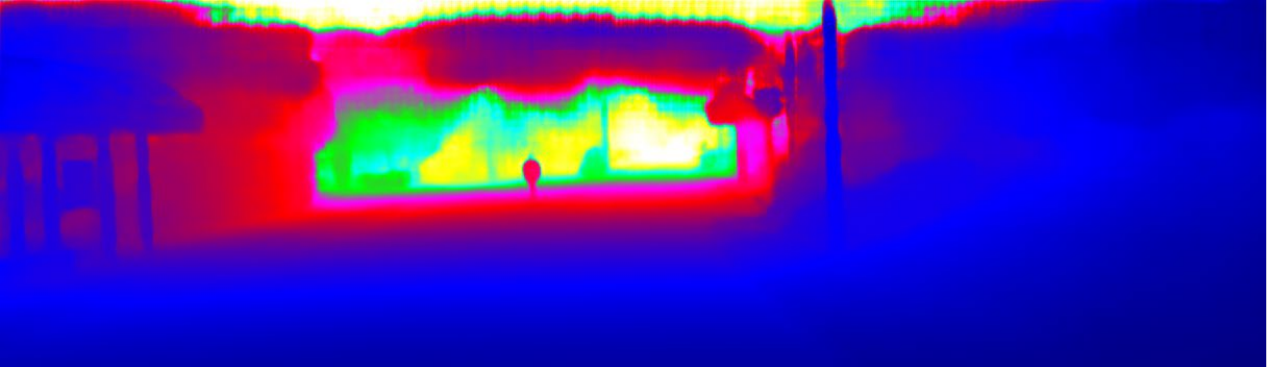}&
       \includegraphics[width=0.30\linewidth]{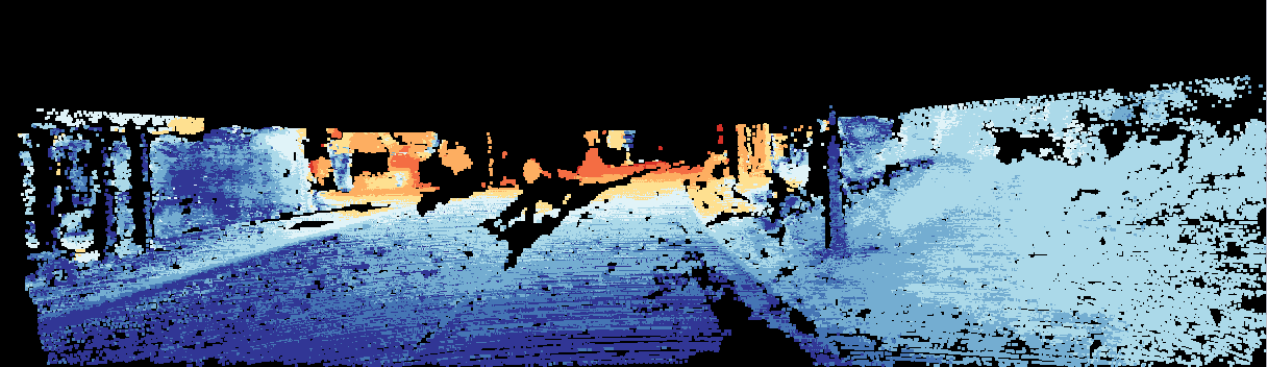}
       \\
       &PWA~\citep{lee2021PWA}&PWA Error~\citep{lee2021PWA}
       \\
       \includegraphics[width=0.30\linewidth]{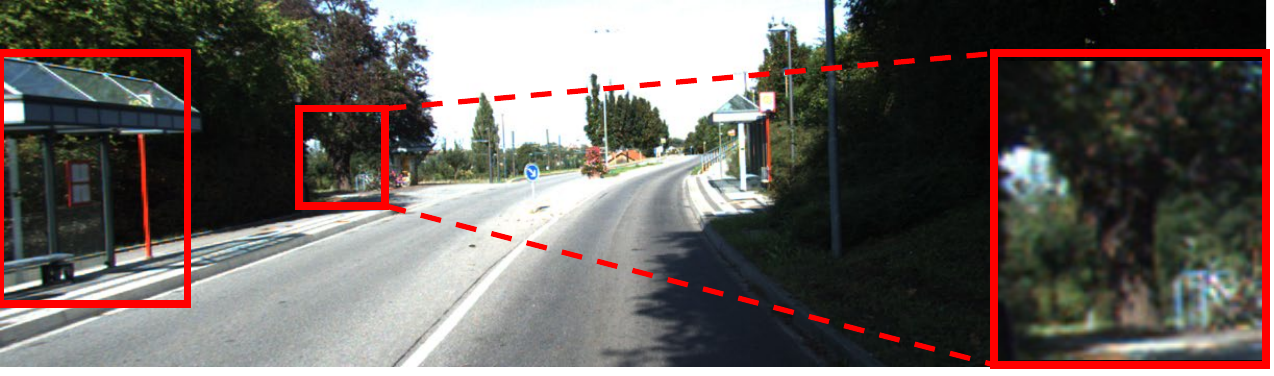}&
       \includegraphics[width=0.30\linewidth]{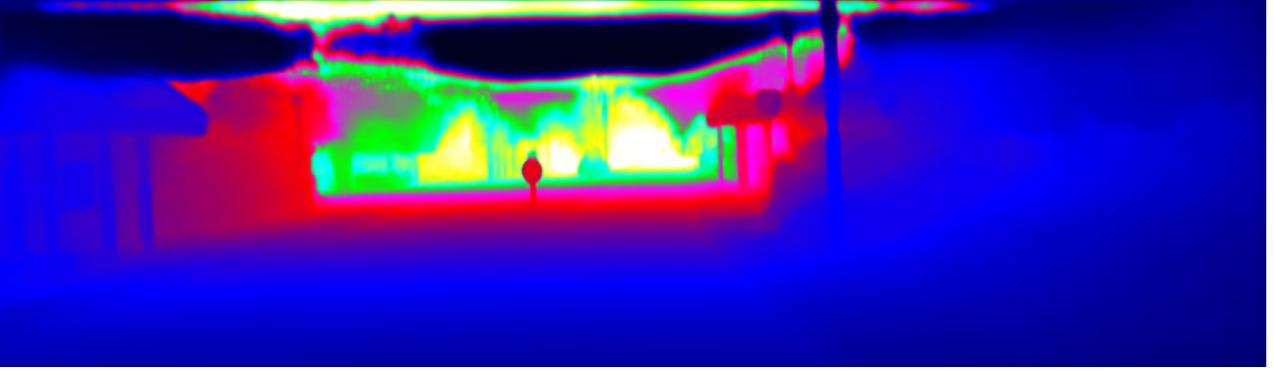}&
       \includegraphics[width=0.30\linewidth]{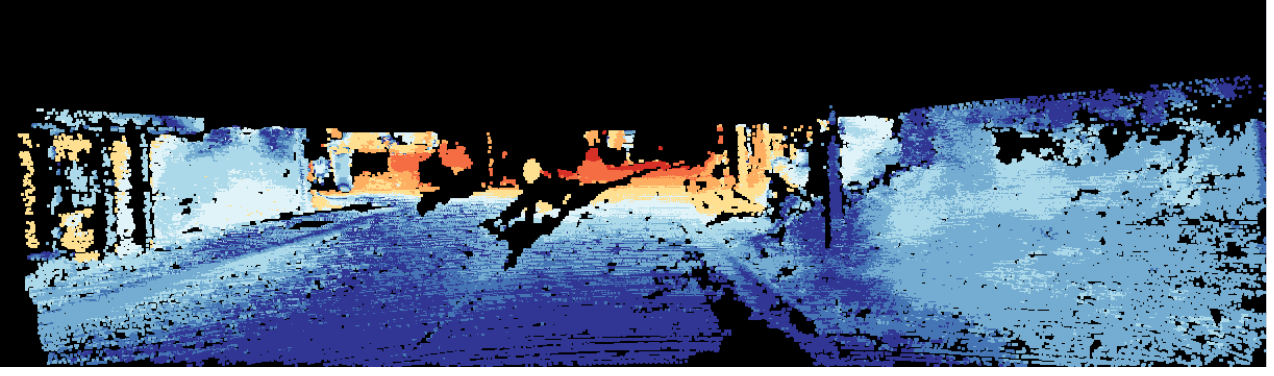}
       \\
       Input&ViP-DeepLab~\citep{qiao2021vip}&ViP-DeepLab Error~\citep{qiao2021vip}
       \\
       &
       \includegraphics[width=0.30\linewidth]{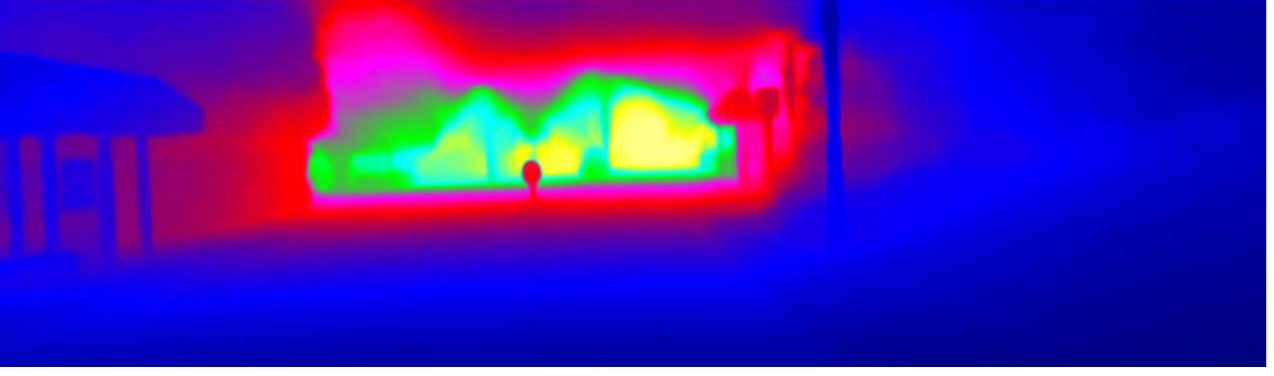}&
       \includegraphics[width=0.30\linewidth]{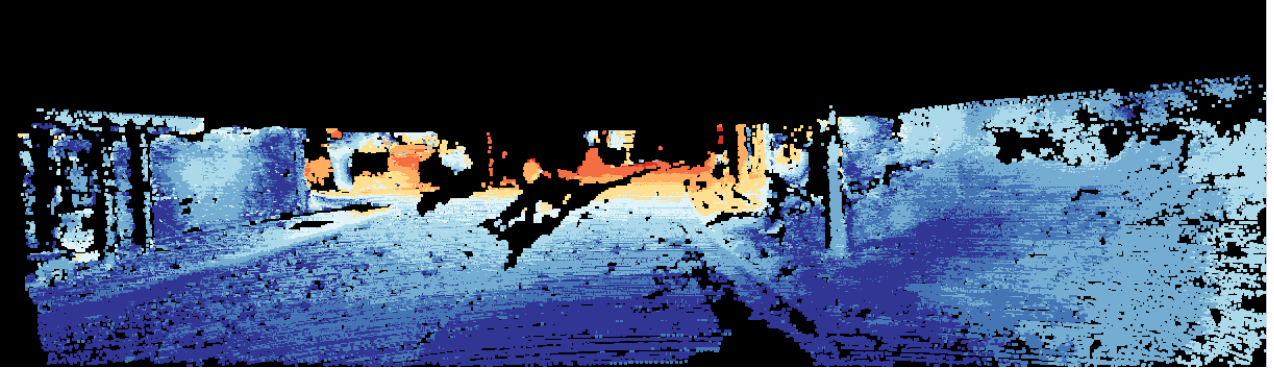}
       \\
       &Ours&Ours Error
       \\
       &
       \includegraphics[width=0.30\linewidth]{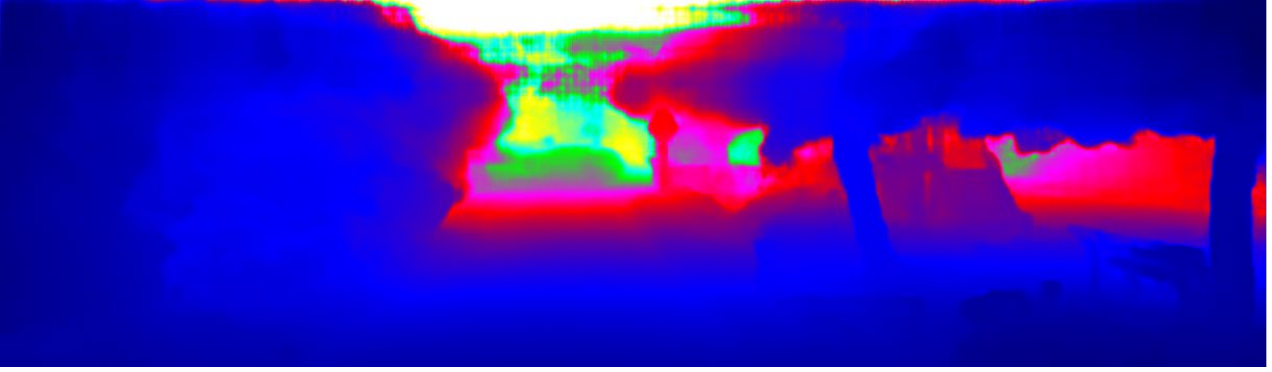}&
       \includegraphics[width=0.30\linewidth]{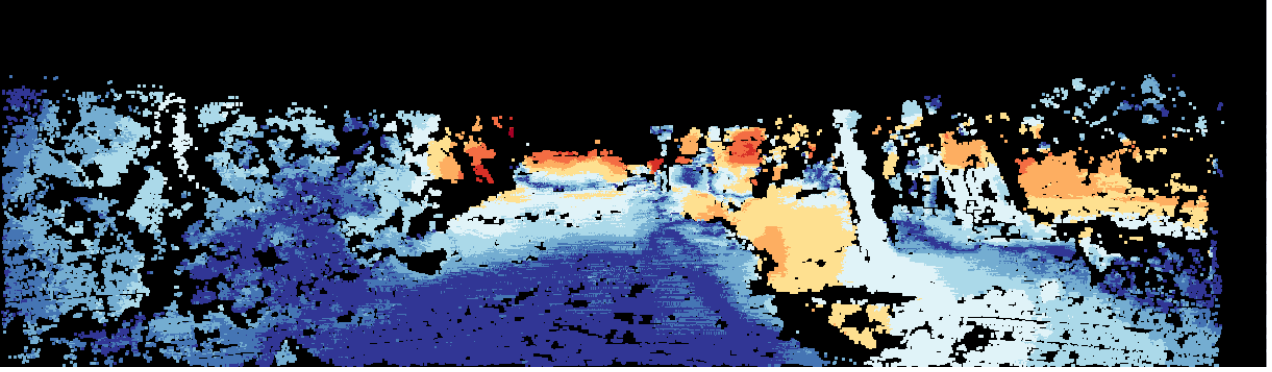}
       \\
       &PWA~\citep{lee2021PWA}&PWA Error~\citep{lee2021PWA}
       \\
       \includegraphics[width=0.30\linewidth]{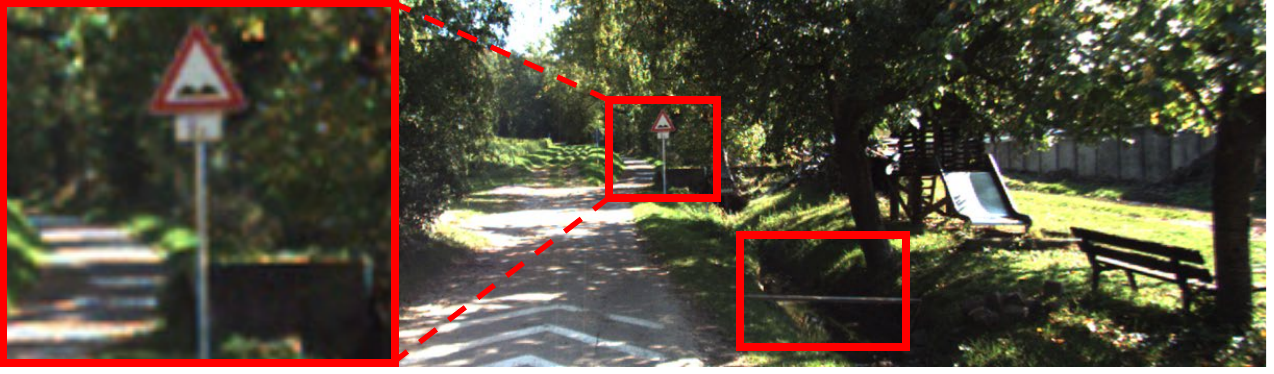}&
       \includegraphics[width=0.30\linewidth]{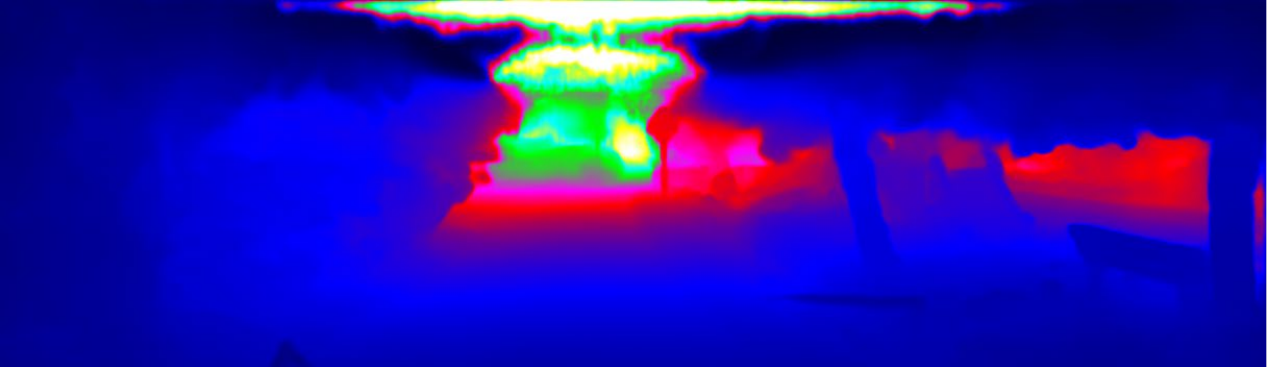}&
       \includegraphics[width=0.30\linewidth]{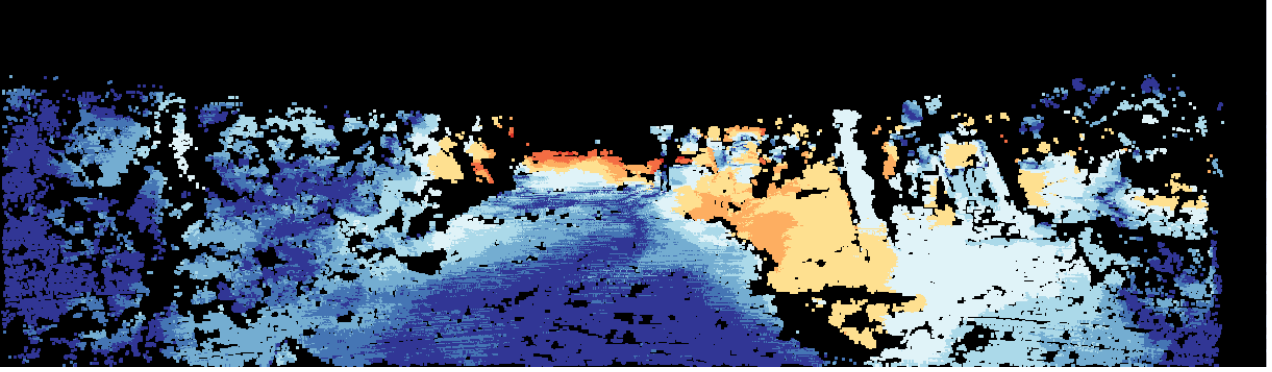}
       \\
       Input&ViP-DeepLab~\citep{qiao2021vip}&ViP-DeepLab Error~\citep{qiao2021vip}
       \\
       &
       \includegraphics[width=0.30\linewidth]{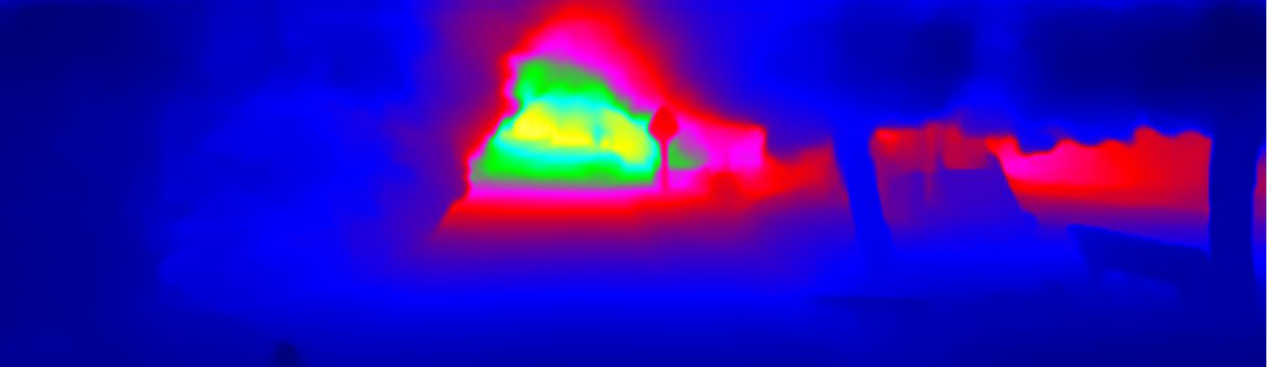}&
       \includegraphics[width=0.30\linewidth]{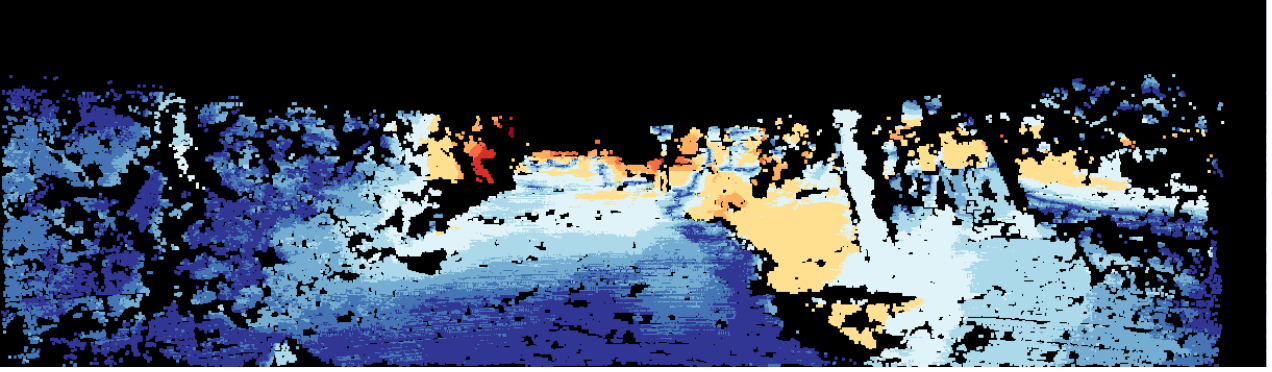}
       \\
       &Ours&Ours Error
       \\
       &
       \includegraphics[width=0.30\linewidth]{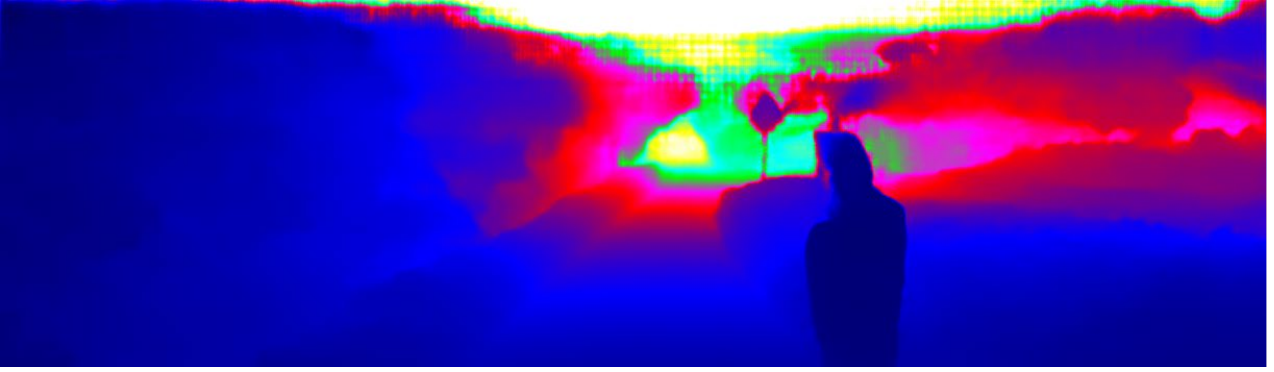}&
       \includegraphics[width=0.30\linewidth]{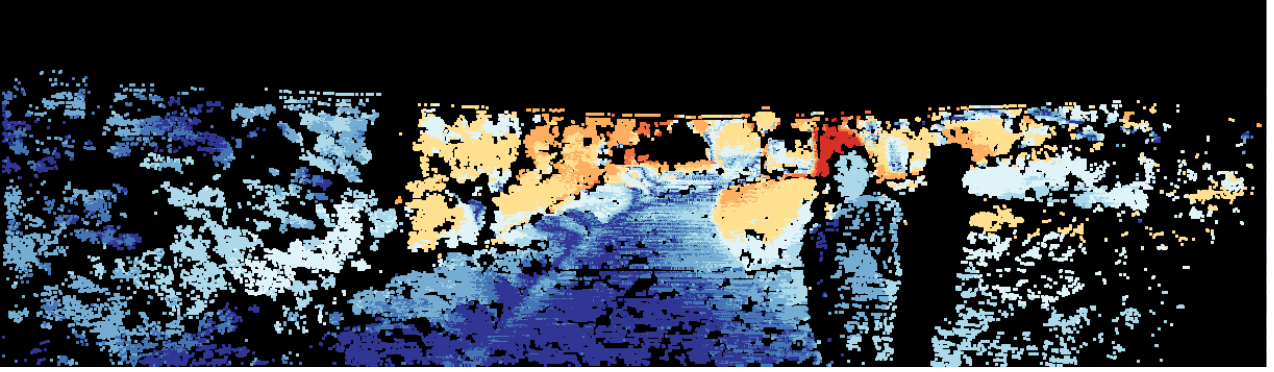}
       \\
       &PWA~\citep{lee2021PWA}&PWA Error~\citep{lee2021PWA}
       \\
       \includegraphics[width=0.30\linewidth]{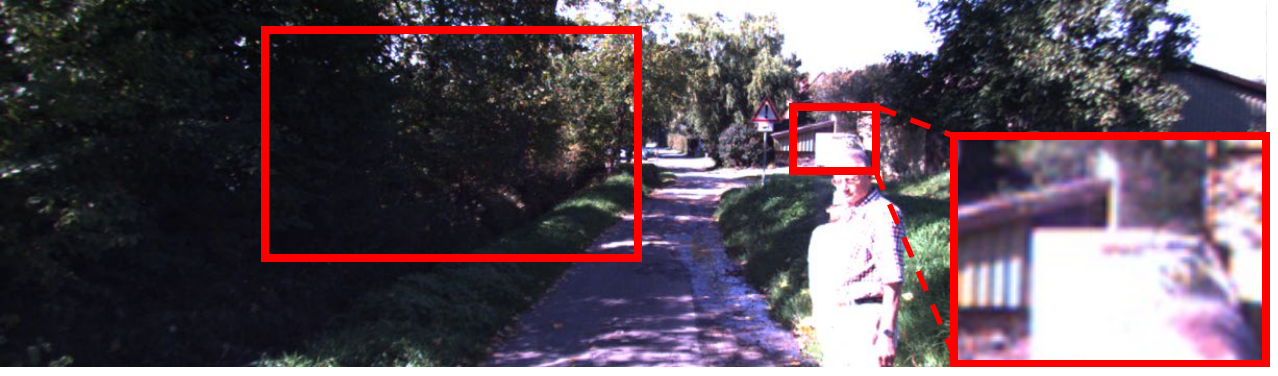}&
       \includegraphics[width=0.30\linewidth]{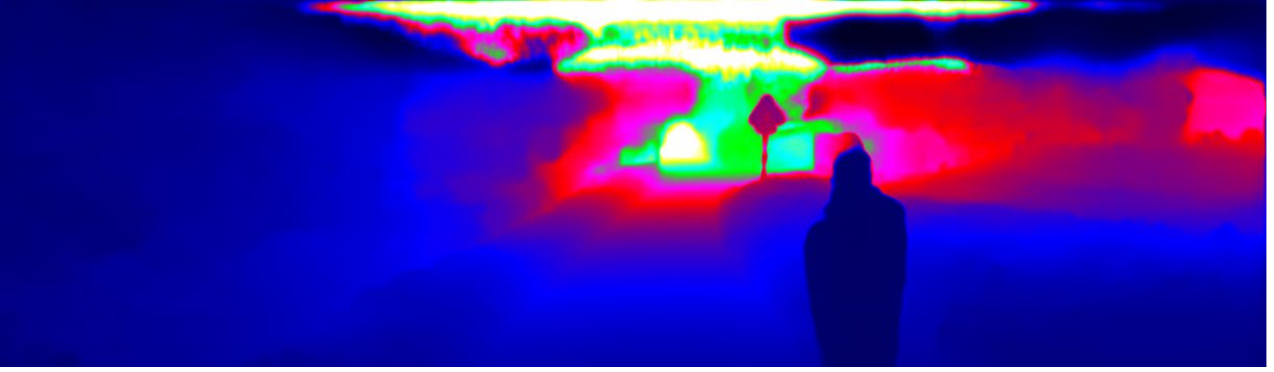}&
       \includegraphics[width=0.30\linewidth]{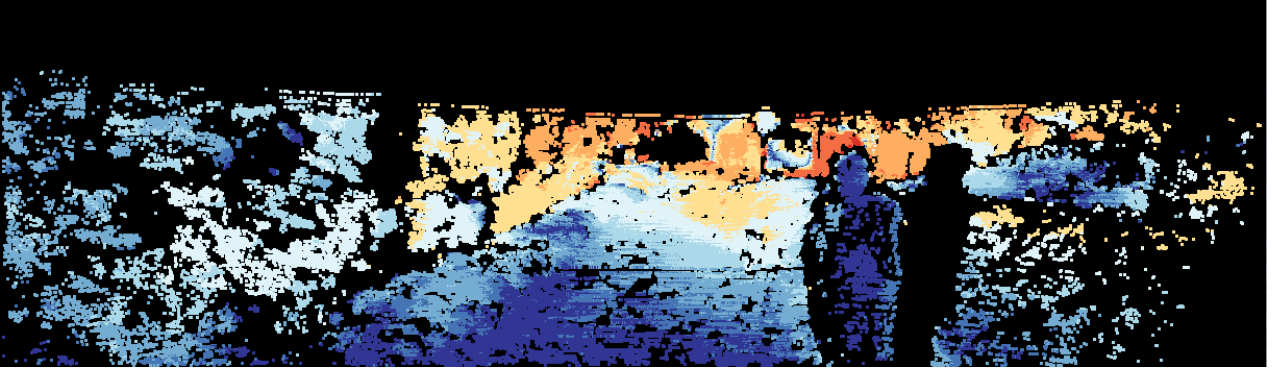}
       \\
       Input&ViP-DeepLab~\citep{qiao2021vip}&ViP-DeepLab Error~\citep{qiao2021vip}
       \\
       &
       \includegraphics[width=0.30\linewidth]{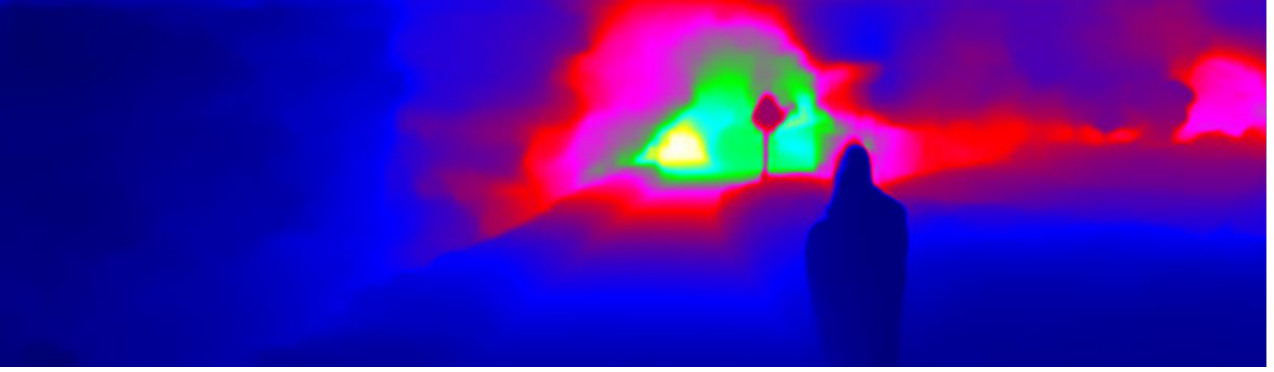}&
       \includegraphics[width=0.30\linewidth]{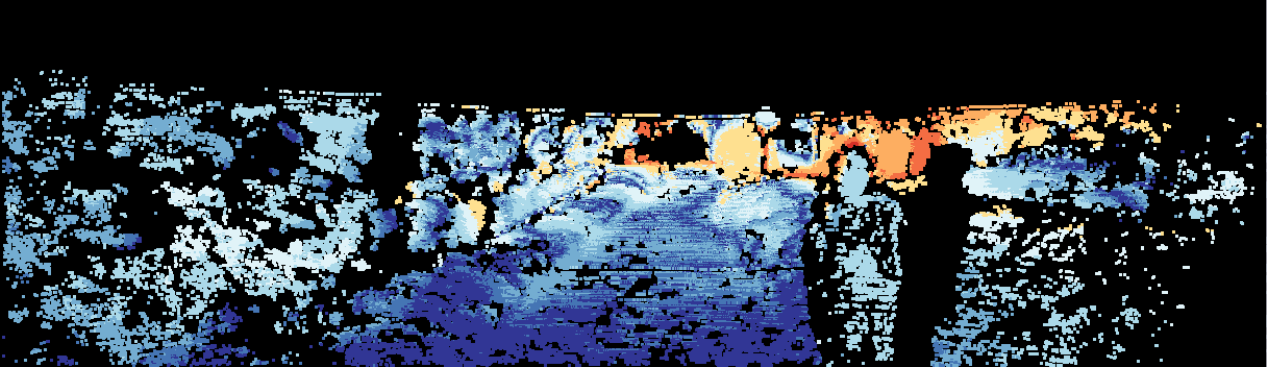}
       \\
       &Ours&Ours Error
   \end{tabular}
   \caption{Qualitative comparison with the state-of-the-art on the KITTI benchmark, better viewed by zooming on screen. Deeper red pixels in the error maps indicate higher errors. Deeper blue means lower errors. The figures are from the official KITTI benchmark website.}
   \label{fig:sup-qualitative-comparison-kitti}
\end{figure*}

\subsection{Comparison to State-of-the-Arts}
We compare the proposed methods with the leading monocular depth estimation models. Primarily, we choose the Adabins~\citep{bhat2021adabins} as our main competitor, which is a solid counterpart and achieved state-of-the-art on all of the datasets we consider. We reproduce the codes of Adabins and load the pre-trained models provided by the authors to get the resulting depth images. Other results are from their official codes.

\textbf{NYU-Depth-v2}: Tab.~\ref{tab:results-nyu} lists the performance comparison results on the NYU-Depth-v2 dataset. While the performance of the state-of-the-art models tends to approach saturation, DepthFormer outperforms all the competitors with prominent margins in all metrics. It indicates the effectiveness of our proposed methods. Qualitative comparisons can be seen in Fig.~\ref{fig:qualitative-comparison-nyu}. DepthFormer achieves more accurate and sharper depth estimation results. We combine camera parameters and predicted depth maps to inv-project the 2D images into the 3D world. As shown in Fig.~\ref{fig:pc}, our reconstructed scenes are satisfying with sharp boundaries of objects and reasonable depth estimations.

\textbf{KITTI}: We evaluate on the Eigen split~\citep{eigen2014depth} and report the results on the Tab.~\ref{tab:results-kitti-val}. DepthFormer significantly outperforms all the leading methods. Qualitative comparisons can be seen in Fig.~\ref{fig:qualitative-comparison-kitti}. We then train our model on the training set of the standard KITTI benchmark split and submit the prediction results of the testing set to the online website. We report the results in Tab.~\ref{tab:results-kitti-test}. While a saturation phenomenon persists in sqErrorRel, DepthFormer still achieves 16\% improvement on this metric and achieves the most competitive result on the highly competitive benchmark as the submission time of Nov. 16th, 2021. We report some qualitative comparison results in Fig.~\ref{fig:sup-qualitative-comparison-kitti}.

\begin{figure*}[t]
   \centering
   \footnotesize
   \begin{tabular}{@{}c@{\hspace{0.1cm}}c@{\hspace{0.1cm}}c@{\hspace{0.1cm}}c@{}}
       \includegraphics[width=0.244\linewidth]{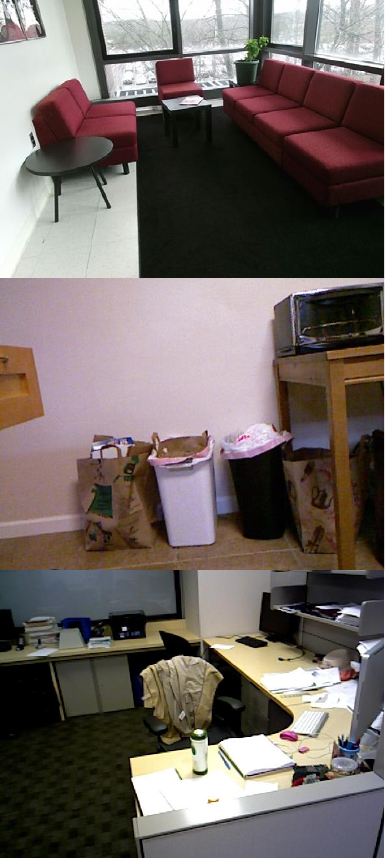} & 
       \includegraphics[width=0.244\linewidth]{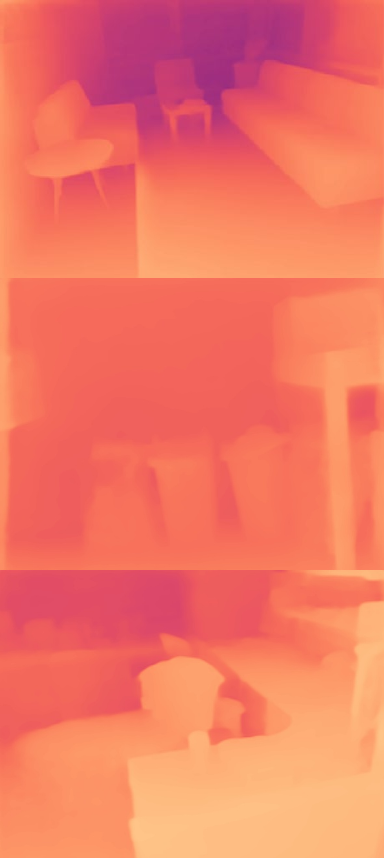} & 
       \includegraphics[width=0.244\linewidth]{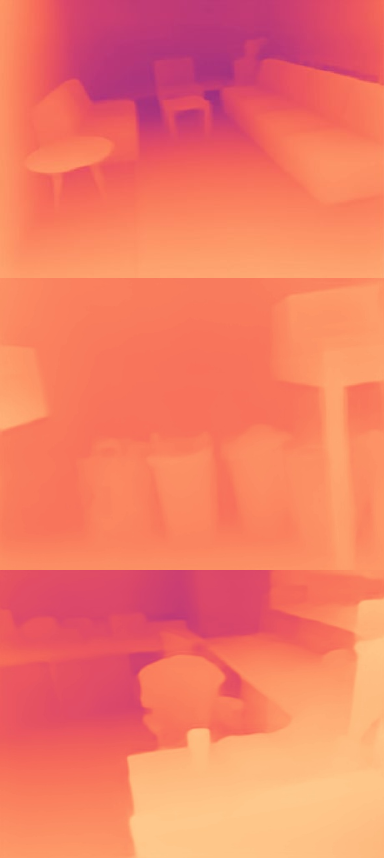} & 
       \includegraphics[width=0.244\linewidth]{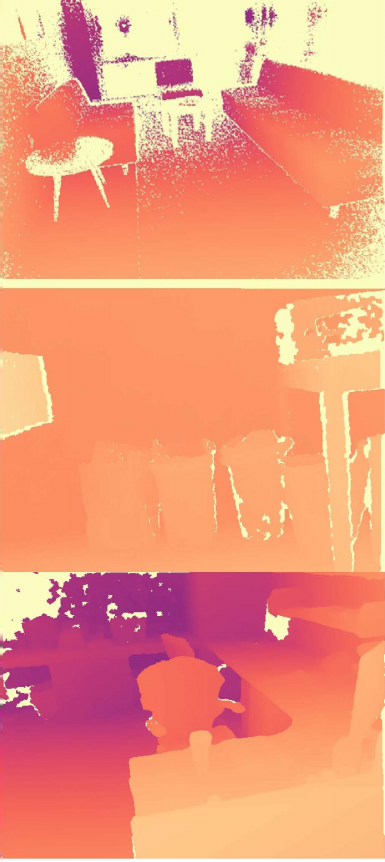} \\
       RGB & Adabins~\citep{bhat2021adabins}& Ours& GT\\
   \end{tabular}
   \caption{Qualitative comparison on the SUN RGB-D dataset.}
   \label{fig:sup-qualitative-comparison-sunrgbd}
\end{figure*}

\textbf{SUN RGB-D}: Following Adabins~\citep{bhat2021adabins}, we conduct a cross-dataset evaluation by training our models on the NYU-Depth-v2 dataset and evaluating them on the test set of the SUN RGB-D dataset without any fine-tuning. As shown in Tab.~\ref{tab:generalization}, significant improvements in all the metrics indicate an outstanding generalization performance of DepthFormer. Qualitative results are shown in Fig.~\ref{fig:sup-qualitative-comparison-sunrgbd}. It is engaging that DepthFormer presents a strong generalization when cross-dataset evaluation. Especially, our method can predict accurate depth estimation for extremely dark areas which are extremely hard to handle without training on the corresponding dataset.

\subsection{Ablation Studies}
For our ablation study, we conduct evaluations with each component of DepthFormer to prove the effectiveness of our method on the NYU and KITTI dataset.

\textbf{Effectiveness of key components}:
We first validate the effectiveness of the key components of DepthFormer. From the baseline network (\textit{i.e.}, ResNet-50, Swin-T), we reinforce the network with our proposed methods and evaluate the improvement of the model performance. The results are reported in the Tab.~\ref{tab:nyu_ablation}. As the additional convolution branch and the HAHI are adopted, the overall performance is significantly improved, which demonstrates the effectiveness of our methods. Moreover, following previous methods~\citep{yang2021transdepth, ranftl2021dpt}, we utilize larger-scale dataset (\textit{i.e.}, ImageNet-22K) to pre-train our encoder. The results (+LP) indicate that the Transformer encoder can better benefit from the larger model capacity and the larger-scale pre-training dataset compared with the CNN encoder.

\begin{figure*}
    \centering
    \includegraphics[width=0.7\linewidth]{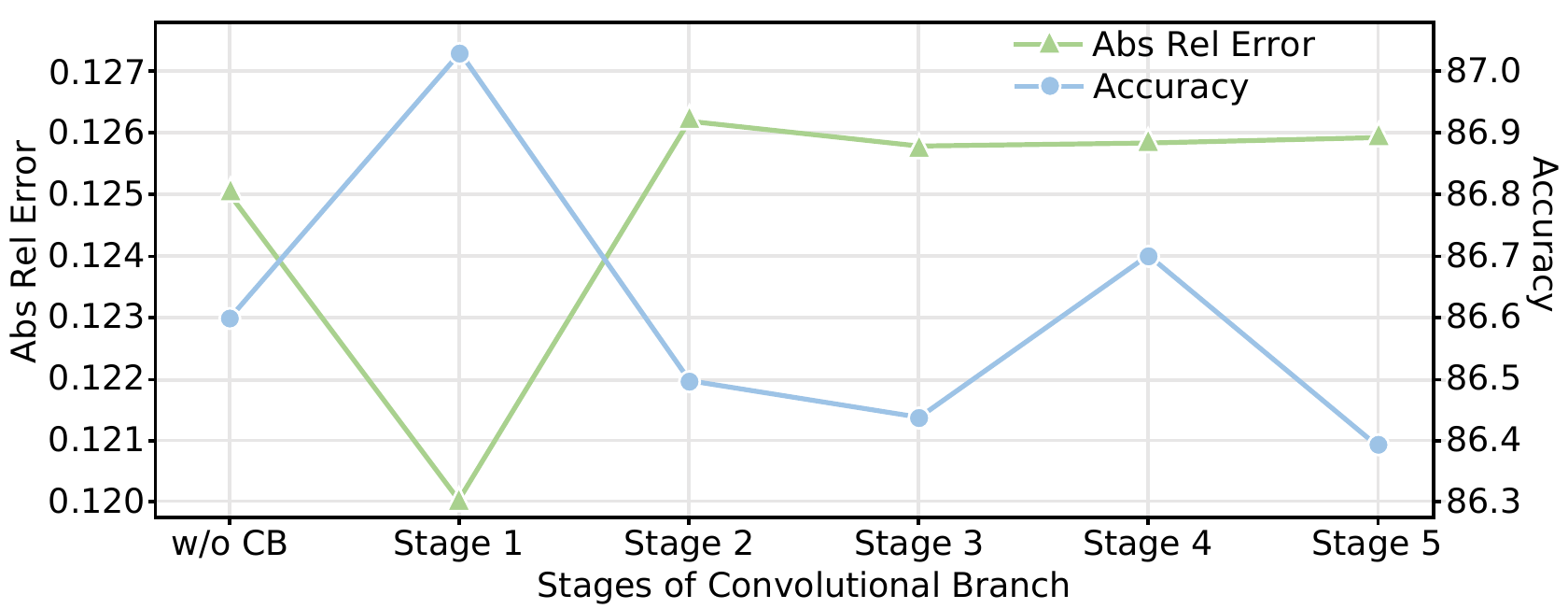}
    \caption{Effect of the convolution branch block on NYU depth estimation performance. We can observe that behaviour decreases after the first block of R-50~\citep{he2016resenet}.}
    \label{fig:ablation-N}
\end{figure*}

\textbf{Fine-grained evaluation on convolution branch}:
The standard CNN encoder can be divided into several sequential blocks. We further scrutinize the influence of different level convolution features on the model performance. Following the default setting, we adopt ResNet-50 as the additional convolution branch. Results are shown in Fig.~\ref{fig:ablation-N}. Interestingly, the model achieves the best performance with only one convolutional block and then downgrades if more blocks are added. A possible explanation for this might be that the consecutive convolutions wash out low-level features, and the gradually reducing spatial resolution discards the fine-grained information~\citep{yang2021transdepth}. Adopting the first block achieves a win-win scenario: it optimizes accuracy by preserving crucial local information while reducing complexity. This can reduce the training time by 2.5$\times$ or more and likewise decrease memory consumption, enabling us to easily scale our Transformer branch to large models.

\textbf{Fine-grained evaluation on HAHI}:
Since the HAHI consists of a deformable self-attention module (DSA) for hierarchical aggregation and a deformable cross-attention module (DCA) for heterogeneous interaction, we conduct more detailed ablation studies on both of these two modules. For fair comparison, we choose the Swin-T with CB as the default backbone. The results are reported in Tab.~\ref{tab:abl_hahi}. We propose to apply the attention mechanism on all the hierarchical features (multi-level DSA) for sufficient aggregation. 
Compared with the one where only each single-layer feature is considered in the attention module, denotes as single-level DSA, the multi-level aggregation strategy get a 4.9\% enhancement on RMS. It demonstrates that the multi-level aggregation strategy is much more effective.
When DSA is added without the multi-level DSA, the model performance is seriously impaired. However, with the multi-level DSA, DCA achieves a 2.2\% improvement on RMS, verifying the importance of both the multi-level DSA and the DCA for heterogeneous interaction. We infer the reason that there are large discrepancies between the heterogeneous features. Multi-level DSA achieves the alignment of the features, which propels the affinity modeling. All the results demonstrate the effectiveness of our proposed HAHI module.

\begin{figure*}
    \centering
    \footnotesize
    \includegraphics[width=0.9\linewidth]{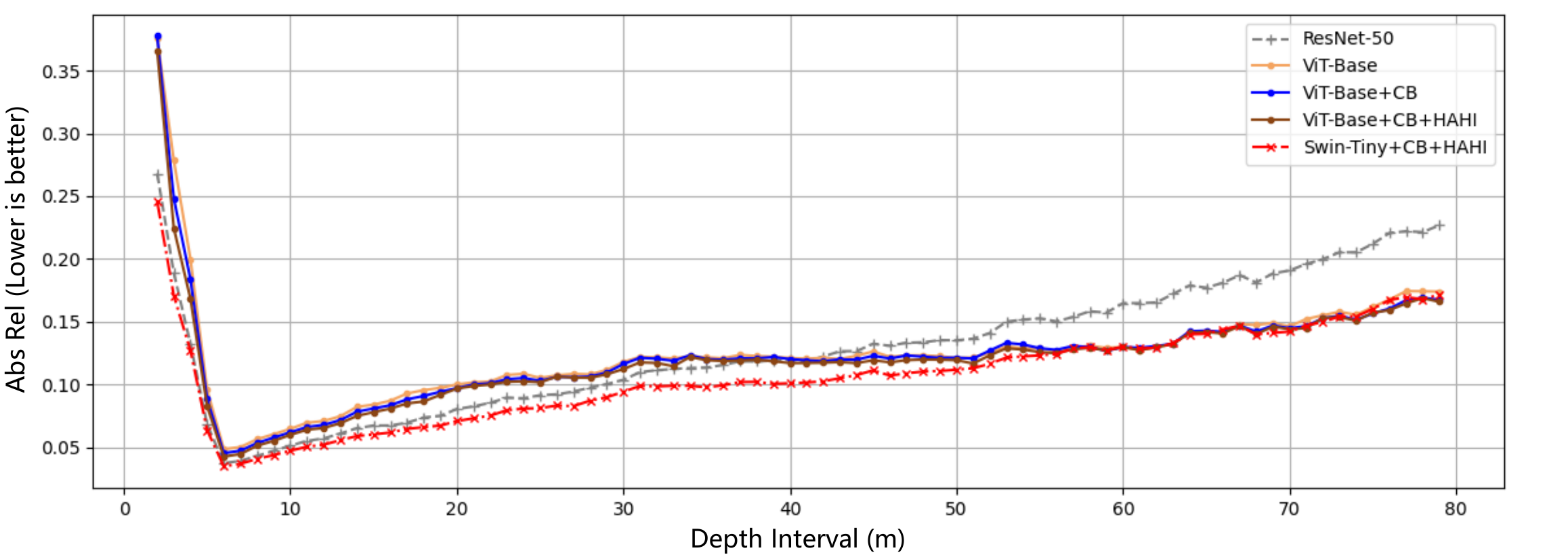}
    \caption{Fine-grained quantitative results of our pilot study on KITTI datset. We divide the depth range (0$m$ - 80$m$) into 80 intervals. Point $(i, j)$ in the plot represents the abs rel of the model is $j$ on depth interval $(i, i+1]m$. Our method achieves a trade off between long and short range estimation.}
    \label{fig:sup-distance}
\end{figure*}

\textbf{Details about pilot study results}:
We have discussed that the CNN branch can provide local information lost in the Transformer branch and the HAHI further promotes the depth estimation via feature enhancement and affinity modeling. They improve the model performance, especially on near object depth estimation. Tab.~\ref{tab:kitti_pilot_extend} demonstrates the effectiveness of our methods. Moreover, we draw more fine-grained results in Fig.~\ref{fig:sup-distance}. Interestingly, Swin-Transformer based model achieves better performance compared with ResNet50-based ones on near object depth estimation and satisfactory results compared with ViT-based ones on distant object depth estimation. We infer that the hierarchical design of the Swin Transformer benefits the extraction of the local information, and the special attention mechanism successfully models the long-range correlation. To compare the model performance in a more direct manner, we also present qualitative comparison results in Fig.~\ref{fig:sup-qualitative-comparison-pilot}. One can observe sharper and more accurate results can be achieved with our proposed CNN branch and HAHI module. 

\begin{figure}
    \centering
    \footnotesize
    \includegraphics[width=1\linewidth]{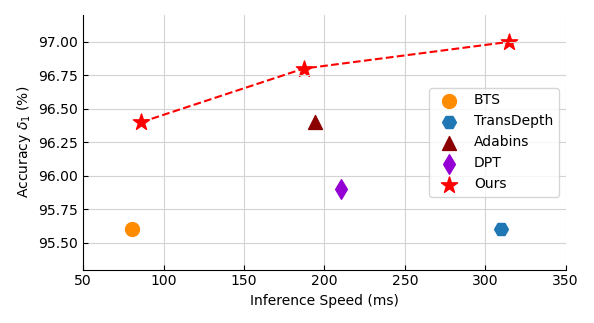}
    \caption{Accuracy $\delta_1$ \textit{vs.} Inference Time on KITTI validation set. The speed is measured using a single RTX 3060 GPU.}
    \label{fig:sup-v}
\end{figure}

\textbf{Inference time evaluation}:
Except for the accuracy ($\delta_1$) of the depth prediction, the inference velocity is of importance as well. We thus evaluate the inference time of DepthFormer w/o HAHI on the KITTI validation set. The resolution of the input images is 352$\times$1216.  Results shown in Fig.~\ref{fig:sup-v} demonstrate that DepthFormer outperforms current state-of-the-art methods in terms of both accuracy and speed.

\section{Conclusion}
\label{sec:conclusion}
We have presented DepthFormer, a novel framework for accurate monocular depth estimation. Our method fully exploits the long-range correlations and the local information by an encoder consisting of a Transformer branch and a CNN branch. Since independent branches with late fusion lead to insufficient feature aggregation for the decoder, we propose the hierarchical aggregation and heterogeneous interaction module to enhance the multi-level features and further model the feature affinity. DepthFormer achieves significant improvements compared with state-of-the-arts in the most popular and challenging datasets. We hope our study can encourage more works applying the Transformer architecture in monocular depth estimation and enlighten the framework design of other tasks.

\textbf{Potential impact}: Beyond the direct application of our work for autonomous driving or spatial reconstruction, there are several venues that warrant future investigation. For example, the common dense global attention in Transformer might be sumptuous. In terms of depth estimation, several key points that indicate the scene structure could be enough to provide crucial long-range information. Designing a more dedicated attention mechanism would improve the effectiveness of the Transformer branch. Furthermore, the HAHI is input-agnostic, and including other modalities such as sparse LiDAR would enhance performance and generalization. Finally, due to the lack of theoretical guarantees, future work to improve the applicability of DepthFormer might consider challenges of explainability and transparency.


\bibliographystyle{spbasic}      
{\footnotesize\bibliography{Mendeley}}

\end{document}